%% file: main.tex
\documentclass[acmtog,nonacm]{acmart} 
\acmSubmissionID{393}

\usepackage{booktabs} 
\usepackage{enumitem}

\usepackage{soul}

\usepackage{bbding}

\usepackage{multirow}
\usepackage[ruled]{algorithm2e} 

\SetAlFnt{\small}
\SetAlCapFnt{\small}
\SetAlCapNameFnt{\small}
\SetAlCapHSkip{0pt}

\usepackage[capitalize]{cleveref}
\crefname{section}{Sec.}{Secs.}
\crefname{table}{Tab.}{Tabs.}
\acmJournal{TOG}

\input{symbols}
\begin{document}
\title{Towards Unified 3D Hair Reconstruction from Single-View Portraits}

\author{Yujian Zheng}
\orcid{0000-0001-7784-8323}
\affiliation{%
 \institution{FNii and SSE, The Chinese University of Hong Kong, Shenzhen}
 \country{China}
}
\email{yujianzheng@link.cuhk.edu.cn}

\author{Yuda Qiu}
\orcid{0009-0003-1257-4271}
\email{yudaqiu@link.cuhk.edu.cn}
\author{Leyang Jin}
\orcid{0009-0002-1440-9096}
\email{leyangjin1@link.cuhk.edu.cn}
\affiliation{%
 \institution{SSE, The Chinese University of Hong Kong, Shenzhen}
 \country{China}
}

\author{Chongyang Ma}
\orcid{0000-0002-8243-9513}
\email{chongyangm@gmail.com}
\author{Haibin Huang}
\email{jackiehuanghaibin@gmail.com}
\author{Di Zhang}
\email{zhangdi08@kuaishou.com}
\author{Pengfei Wan}
\email{wanpengfei@kuaishou.com}
\affiliation{%
 \institution{Kuaishou Technology}
 \country{China}
}

\author{Xiaoguang Han}
\authornote{Corresponding author: Xiaoguang Han (hanxiaoguang@cuhk.edu.cn).}
\orcid{0000-0003-0162-3296}
\affiliation{%
 \institution{SSE and FNii, The Chinese University of Hong Kong, Shenzhen}
 \country{China}}
\email{hanxiaoguang@cuhk.edu.cn}

\input{fig/teaser}
\begin{abstract}

Single-view 3D hair reconstruction is challenging, due to the wide range of shape variations among diverse hairstyles.
Current state-of-the-art methods are specialized in recovering un-braided 3D hairs and often take braided styles as their failure cases, because of the inherent difficulty to define priors for complex hairstyles, whether rule-based or data-based. We propose a novel strategy to enable single-view 3D reconstruction for a variety of hair types via a unified pipeline. 
To achieve this, we first collect a large-scale synthetic multi-view hair dataset \dataset with diverse 3D hair in both braided and un-braided styles, and learn two diffusion priors specialized on hair.
Then we optimize 3D Gaussian-based hair from the priors with two specially designed modules, i.e. view-wise and pixel-wise Gaussian refinement.
Our experiments demonstrate that reconstructing braided and un-braided 3D hair from single-view images via a unified approach is possible and our method achieves the state-of-the-art performance in recovering complex hairstyles.
It is worth to mention that our method shows good generalization ability to real images, although it learns hair priors from synthetic data.
Code and data are available at \href{https://unihair24.github.io/}{https://unihair24.github.io}

\end{abstract}

%
%


%
%


\maketitle

\input{sec/introduction}
\input{fig/pipeline}
\input{sec/related_work}

\input{sec/overview}

\input{sec/dataset}
\input{sec/methods}
\input{sec/experiments}
\input{sec/conclusion}


\bibliographystyle{ACM-Reference-Format}
\bibliography{main}




\appendix 
\renewcommand{\appendixname}{Supplementary Material~\Alph{section}}

\setcounter{footnote}{0}
\setcounter{table}{0}
\setcounter{figure}{0}
\renewcommand\thesection{\Alph{section}}
\renewcommand{\thetable}{S\arabic{table}}  
\renewcommand{\thefigure}{S\arabic{figure}}

\vspace{3em}
\centerline{\textbf{\LARGE{-- {Supplementary Material} --}}}

\input{sec/appendix}
\clearpage
\end{document}

%% file: symbols.tex
\newcommand{\dataset}{\emph{SynMvHair}\xspace}

\newcommand{\synthesizer}{\emph{HairSynthesizer}\xspace}
\newcommand{\deblurer}{\emph{HairEnhancer}\xspace}

\newcommand{\inputimage}{$I$\xspace}
\newcommand{\alignedimage}{$I_a$\xspace}
\newcommand{\coarsegs}{$\Theta^0$\xspace}
\newcommand{\refinedgs}{$\Theta^1$\xspace}
\newcommand{\enhancedgs}{$\Theta^2$\xspace}

%% file: fig/teaser.tex
\begin{teaserfigure}
\centering
  \includegraphics[width=1.0\textwidth]{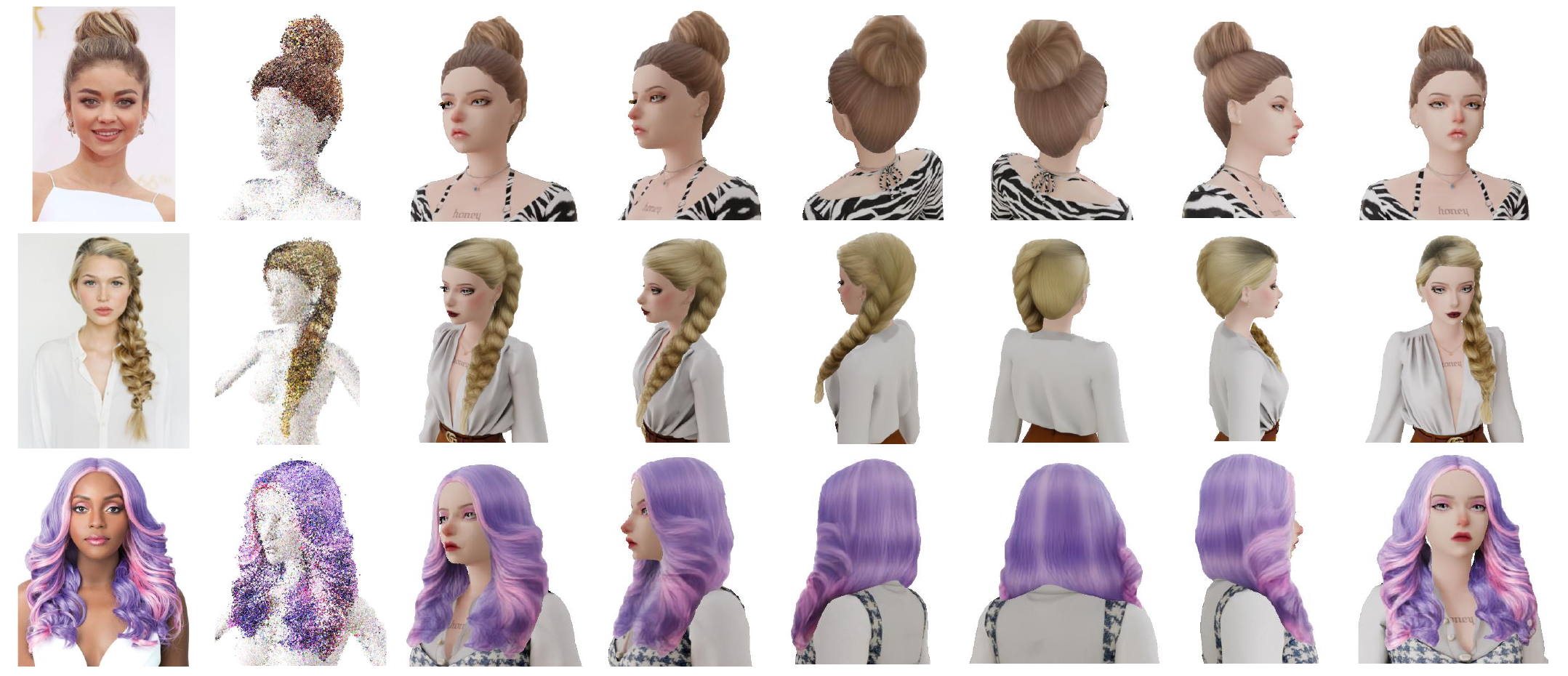}
  \caption{We proposed a novel strategy which enables single-view 3D hair reconstruction with rich textures and complicated shapes, working for braided and un-braided hairstyles through a unified pipeline. From left to right: input image, reconstructed 3D hair in 3D Gaussian representation and its rendered images from multiple views. Note that, our reconstruction is only for hair part and the underlying virtual avatar is just for visualization. }
  \label{fig:teaser}
\end{teaserfigure}

%% file: sec/introduction.tex
\section{Introduction}
\label{sec:introducation}

Human hair varies in a wide range of styles, from short but classical Bob cut to princess-style long hair with fancy braids, buns or twists. 
The large shape variations among such diverse hairstyles make single-view 3D hair reconstruction a challenging task.

Modeling multiple hair types in a single framework is desirable, but currently disparate techniques often are used for disparate hairstyles. One important example we aim to solve is handling both braided and un-braided hair in the same system.
Following early 2D strand lifting methods~\cite{chai2012single,chai2013dynamic,chai2015high}, \cite{sun2021single} reconstructs the braided style via specifically designed braid unit identification, but fails to generate non-visible parts of the hair.
State-of-the-art methods use retrieval-based deformation~\cite{hu2015single,chai2016autohair,hu2017avatar} or fully-supervised learning approaches~\cite{zheng2023hairstep,wu2022neuralhdhair,zhou2018hairnet} to reconstruct 3D hair in the representation of strands or strips. 
They are limited by the small scale and narrow style range of 3D hair datasets~\cite{hu2015single,chai2016autohair} and can only deal with simple un-braided hairstyles.
However, creating 3D hair strands requires substantial human effort, especially for complex styles like braids and buns with inner weaved geometries.

Inspired by the success of generative image-to-3D approaches~\cite{liu2023zero,tang2023dreamgaussian}, 3D object reconstruction can also be realized using optimization based 2D-lifting methods relying on 2D diffusion priors trained on rendering-only 3D dataset~\cite{deitke2023objaverse}, not requiring high-quality full geometries.
However, it is not straightforward to apply existing generative image-to-3D methods~\cite{liu2023zero,tang2023dreamgaussian} to single-view 3D hair reconstruction. 
There are two main issues unresolved.
Firstly, current priors are learned from images of general objects rather than being specifically tailored to the hair domain. This often results in suboptimal hair reconstruction outcomes. 
Additionally, there is no available large-scale 3D hair dataset that covers a sufficient range of hairstyles or has the scale necessary to train satisfactory priors.
Secondly, compared to the texture of common objects, hair texture presents more challenges. It demands high-quality, strand-like details. However, current approaches can only achieve limited quality when dealing with such high-frequency hair textures.

In this work, we introduce a unified strategy that first enables the reconstruction of diverse 3D hairstyles from single-view images.
It is a coarse-to-fine optimization-based 2D lifting method, which gets rid of the dependence on expensive 3D hair data and turn to optimize 3D hair from 2D diffusion priors.
We choose the popular 3D Gaussian~\cite{kerbl3Dgaussians} as the underlying 3D hair representation. 3D Gaussian is easy to achieve realistic rendering and has good flexibility. Also, the optimization of 3D Gaussian is proved fast.
To obtain satisfactory hair priors, we build a large-scale dataset \dataset of calibrated multi-view hair images covering diverse hairstyles.
Based on \dataset, we prepare two diffusion-based hair priors, \synthesizer for synthesizing novel views from single-view input and \deblurer for removing the blurry pattern of hair texture, which empower to distill fine 3D hairs conditioning on single-view images. Specifically, \synthesizer has similar architecture as Zero-1-to-3~\cite{liu2023zero}, which generates images of target views from random noises conditioned on given images and relative camera poses. Taking a blurry image as condition, \deblurer can generate the image with high-quality texture and maintain good alignment with the input in pixel-level.

In the process of Gaussian optimization, we observe that directly applying Score Distillation Sampling (SDS) loss~\cite{poole2022dreamfusion} against \synthesizer leads to noisy and blurry hair texture. 
To reconstruct 3D hair with fine-grained textures, we carefully design two different level Gaussian refinements. 
We first perform a view-level Gaussian refinement with the supervision of dense views generated by \synthesizer.
Although the noise can be clearly removed, the texture is still blurry due to the inherent view-inconsistency of \synthesizer.
Thus, we further conduct a pixel-level Gaussian refinement against \deblurer and obtain a satisfactory 3D hair with fine texture, as shown in~\cref{fig:teaser}.

Extensive experiments have demonstrated the efficacy of our approach on single-view 3D hair reconstruction across diverse hairstyles.
It is noted that our output, Gaussian-based 3D hair, can achieve fast, high-quality, and consistent multi-view hair rendering.
Thus, an extra bonus of our method is to convert single-view 3D strands reconstruction to the relatively maturer multi-view strands capturing~\cite{sklyarova2023neural,hu2014capturing}.
Compared with state-of-the-art single-view 3D strand reconstruction method~\cite{zheng2023hairstep}, combining our method and NeuralHairCut~\cite{sklyarova2023neural} achieves more reasonable results on un-braided hairstyles, especially for the invisible region from the given view.

The main contributions of this work are listed as follows:
\begin{itemize}[leftmargin=*]
\item We propose a novel strategy to reconstruct textured 3D hair from single-view images, covering diverse hairstyles within a single pipeline. Our approach handles both simple un-braided styles, which are covered by state-of-the-art methods~\cite{zheng2023hairstep,wu2022neuralhdhair}, as well as more complex braided styles such as ponytails, French braids, and buns.
\item We contribute a large-scale synthetic hair dataset \dataset, covering various hairstyles, of calibrated multi-view images related to 2,396 collected 3D hair and 82,682 texture maps in total, which facilitates the learning of hair diffusion priors. It will be released to benefit the hair research.
\item We carefully design a Gaussian-based single-view 3D hair reconstruction method, which adopts a coarse-to-fine manner, including both view-level and pixel-level Gaussian refinements, to achieve high-quality and view-consistent results.
\item Our method demonstrates state-of-the-art performance in recovering complex hairstyles from single-view inputs and exhibits strong generalization capabilities on real portraits, even though it learns hair priors solely from synthetic data.
\end{itemize}

%% file: fig/pipeline.tex
\begin{figure*}
\centering
\includegraphics[width=1.0\linewidth]{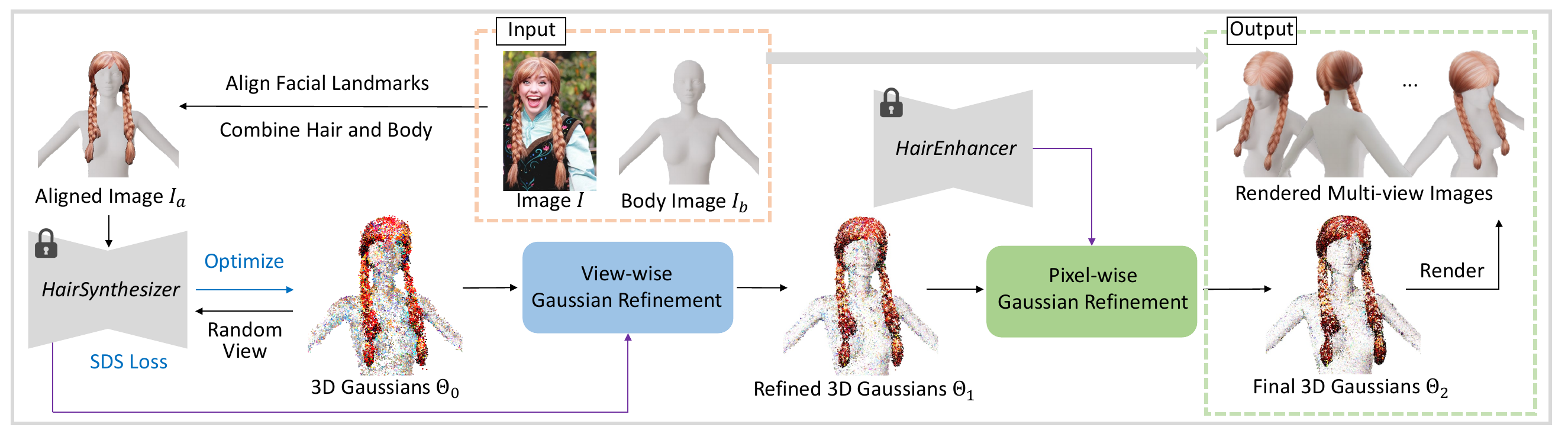}
\caption{Overview of our method. Given only a single-view portrait, our method can reconstruct its corresponding 3D hair in the representation of 3D Gaussian and render high-quality images from arbitrary views.}
\label{fig:pipeline}
\end{figure*}

%% file: sec/related_work.tex
\section{Related Work}
\label{sec:related_work}

\paragraph{{Single-view 3D hair reconstruction.}}
Reconstructing high-fidelity 3D hairstyles from a single-view input has been acknowledged as a challenging problem in creating digital human. 
The pioneering works~\cite{chai2012single, chai2013dynamic} obtain 3D hair models by tracing and lifting 2D curve according to given user-input stokes. 
These methods fail to recovery plausible geometry of the invisible parts cause by occlusion. 
To tackle this problem, retrieval-based approaches~\cite{hu2015single, chai2016autohair, hu2017avatar} introduce 3D hairstyle priors from synthetic dataset and achieve better quality. 
However, given limited searching efficiency, the size and diversity of the database has been the bottleneck of their performances. 
With the booming of deep leaning, several methods based on CNN~\cite{zhou2018hairnet} and generative models~\cite{saito20183d, zhang2019hair} have been proposed. 
Mainly due to the limited capacity of 3D representation, the results of these methods tend to be coarse and over-smoothed. 
Recently, methods that represent hair modeling by strand growing according to implicit fields~\cite{yang2019dynamic, wu2022neuralhdhair, zheng2023hairstep} have achieved the state-of-the-art performances. However, these methods fail to depict some special growing pattern like braids or hair buns and shows unsatisfactory generalization ability, because their depending datasets only covers a small scale un-braided hairstyles. 
Following early 2D strand lifting methods~\cite{chai2012single,chai2013dynamic,chai2015high}, \cite{sun2021single} focuses on reconstructing the braided style via braid unit identification, but still fails to generate non-visible parts of the hair.
Currently, there is no single-view 3D hair reconstruction method covering un-braided and braided styles in a unified pipeline. 

\paragraph{{3D hair capture.}}
3D hair capture with more input, like multi-view images and RGBD images, is maturer and easier than single-view modeling. 
Multi-view input provides an ideal condition for high-quality hair capture~\cite{luo2012multi, luo2013wide, luo2013structure, hu2014robust, nam2019strand, rosu2022neural, sklyarova2023neural, wu2024monohair, luo2024gaussianhair} and shows good reconstruction quality.
To ease the harsh scenarios of dense camera array and large poses, some methods are proposed with input of sparse views~\cite{zhang2017data, kuang2022deepmvshair}, selfie videos~\cite{liang2018video} and sparse RGBD streams~\cite{zhang2018modeling}. 
\cite{shen2023ct2hair} provides a method to recover 3D hair strands from CT scans, but it is not possible to reach common consumers.
All these methods struggles when attempting to recover braided styles.
\cite{hu2014capturing} focuses on capturing several braided hairstyles with dense RGBD inputs with an example database.
Despite of the requirement of specialized equipment or computation power for easing the input from single-view image to multi-view or other types, reconstructing braided and un-braided 3D hair in a unified pipeline has still not been achieved yet.

\paragraph{{Image-to-3D}} Facing 3D data shortage, another possible way for single-view 3D reconstruction is to make full use of 2D information which is easier to be obtained. 
Based on diffusion models~\cite{liu2023zero,radford2021learning} learned from large-scale 2D datasets, existing Image-to-3D methods using optimization-based 2D lifting to reconstruct 3D objects in the representation of textured mesh~\cite{liu2023one}, NeRF~\cite{liu2023zero} or Gaussian~\cite{tang2023dreamgaussian}.
The good generalization ability of these methods show the possibility for their application on recovering general 3D hair model with overall shape and texture.
However, it is not straightforward because of the unique domain of hairstyles and the challenging requirement of high-quality hair texture.

%% file: sec/overview.tex
\section{Overview}
\label{sec:overview}

As shown in~\cref{fig:pipeline}, given an input portrait image, our method first enables the reconstruction of its corresponding 3D hair in the representation of 3D Gaussian.
Our method supports input of various hairstyles, ranging from un-braided Bobs and Waves to knotted braids and buns, etc.
The obtained 3D hair is aligned with a template body and can be rendered into high-quality images from arbitrary views.
To support our method, we construct a novel dataset \dataset and build two diffusion priors \synthesizer and \deblurer.
We describe our \dataset in~\cref{sec:dataset} and how the pipeline works in~\cref{sec:method}.

%% file: sec/dataset.tex
\section{Dataset}
\label{sec:dataset}

\input{fig/dataset}
We construct the \dataset dataset to support a large-scale data-driven approach for reconstructing diverse 3D hair varying in braided and un-braided styles.
\dataset contains 2,396 3D hair shapes and 82,682 textures in total, where 1,544 un-braided and 852 braided hairstyles are included and each hair shape has 5-188 different UV texture maps.

We first collect 10,320 different raw synthetic hair models from The Sims Resource\footnote{\url{{https://www.thesimsresource.com/}}}. 
After taking efforts to check and remove the poor data with coarse shape and blurry texture, we finally obtain 2,396 3D hair models and 82,682 textures in total.
All of these 3D hair models are composed of hundreds to thousands independent thin polygon-strips and each strip represents a coherent hair wisp. 
Furthermore, all 3D hair models are canonical and normalized by artists, in other words, they are aligned onto the head of an identical template body.
As the inner structure of 3D hairs are not as good as desired, we cannot convert them to 3D strands or strips connected with hair roots within an allowable budget.
Inspired by~\cite{liu2023zero}, we render dense multi-view images with high-quality hair textures to enable the learning of 3D-aware diffusion priors. 
The camera can randomly move on a upper hemisphere whose center is roughly set at the center of neck and whose radius is set to contain the upper-half body for shooting most of hairstyles.
A data sample of \dataset is illustrated in~\cref{fig:dataset}.

\paragraph{{Comparisons with existing dataset.}}
\cite{hu2015single} built the first synthetic 3D hair dataset with the scale of 343 un-braided hairstyles.
\cite{chai2016autohair} claimed a same dataset with 653 un-braided models, but it is not public available.
On the contrast, we provide 2,396 3D hair models with 1,544 un-braided and 852 braided styles, which expands the current 3D hair datasets on both the scale and the style, especially for braided styles, like single or double ponytails, pigtails, French braids, French twists, Gretchenfrisur, buns, etc.
Compared with existing 2D dataset like~\cite{kim2021k}, our \dataset has normalized and canonical 3D hair models to render multi-view images with calibrated and accurate cameras.
For back and side view of captured images in~\cite{kim2021k}, it is difficult to estimate accurate cameras~\cite{an2023panohead}.
Although \cite{an2023panohead} attempted to address this issue, serious artifacts like Janus problem are still cannot be avoided.
Thus, applying such dataset to enable 3D hair reconstruction is not easy to achieve at present.

%% file: fig/dataset.tex
\begin{figure}
\centering
\includegraphics[width=1.0\linewidth]{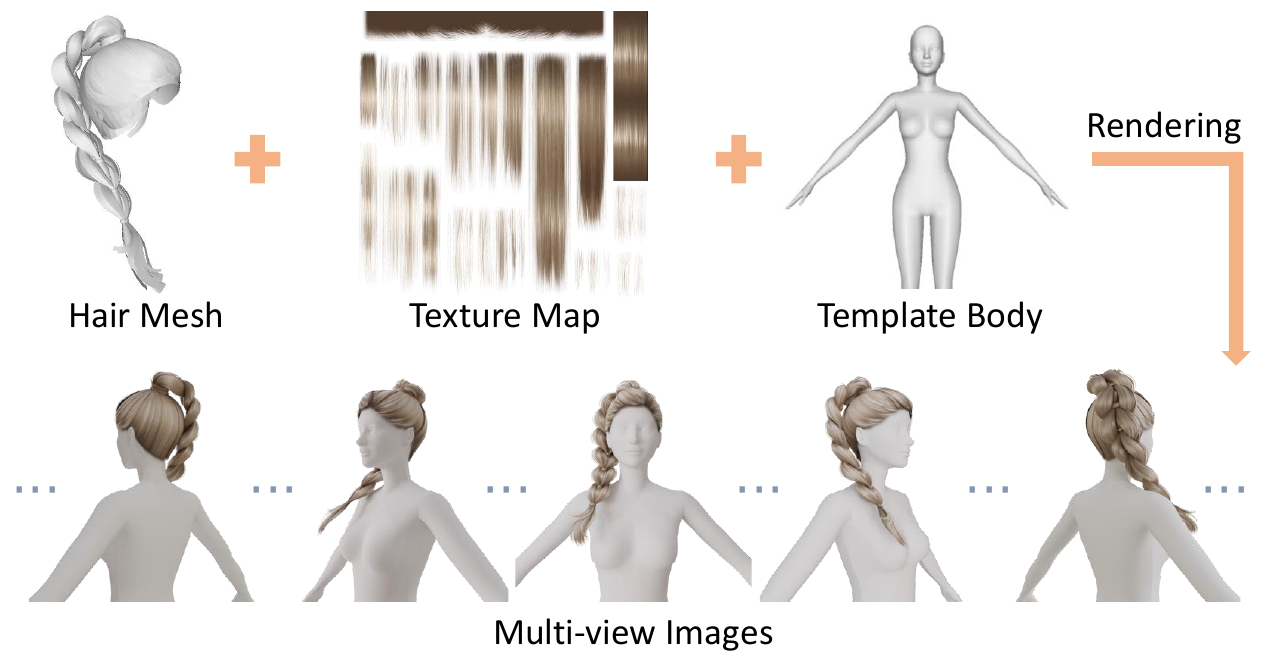}
\caption{Illustration of a sample in our \dataset dataset.}
\label{fig:dataset}
\end{figure}

%% file: sec/methods.tex
\section{Methods}
\label{sec:method}

As shown in Fig.~\ref{fig:pipeline}, we first wears the hair of the input image \inputimage on the rendered template body to align the format of the condition in \synthesizer.
Then a coarse 3D Gaussian \coarsegs will be distilled from \synthesizer via an optimization guided by SDS loss~\cite{poole2022dreamfusion} conditioning on the aligned image \alignedimage.
However, the hair texture provided by \coarsegs is noisy and blurry.
Thus, we further perform view-wise and pixel-wise Gaussian refinements leveraging the prior of \synthesizer and \deblurer, to obtain refined 3D Gaussian \refinedgs and enhanced 3D Gaussian \enhancedgs, respectively.
Thanks to the two diffusion priors based on our novel dataset \dataset and the level-wise refinement module, our method can finally output a Gaussian-based 3D hair which enabling multi-view rendering with fine hair textures.

\subsection{Image Pre-processing}
The whole system is built upon \dataset where hair images are rendered with a same template body.
To reconstruct a 3D hair from a real portrait, the first step is aligning the hair in \inputimage to the rendered frontal image $I_b$ of template body. 
We conduct this process via facial landmark alignment.
We first annotate 68 2D landmarks $lmk_b$ on the face of the body image $I_b$ and detect landmarks $lmk_h$ on \inputimage via 3DDFA$\_$V2~\cite{guo2020towards}.
Then we optimize the translation, rotation and scale $\{t,r,s\}$ for aligning $lmk_h$ to $lmk_b$ via minimizing the $L_2$ distance between corresponding 2D points in two landmark sets.
The hair region $I_h$ in \inputimage will be masked via prompting 'hair' text to the segmentation tool~\cite{kirillov2023segany,liu2023grounding,ren2024grounded}.
Then the aligned image \alignedimage can be formed after wearing $I_h$ to $I_b$ using $\{t,r,s\}$.
Please note that all images in this work are at the resolution of $512\times512$, and the real image will be padded and resized before inputted.

\subsection{Coarse 3D Gaussian Optimization}
To reconstructing 3D hair shown in \alignedimage using \dataset, we follow~\cite{liu2023zero} to build a diffusion-based module \synthesizer for novel view synthesis of hair images.
Although \synthesizer can generate plausible novel views of a given hair image, the consistency between different views cannot be guaranteed.
Thus, based on \synthesizer, we optimize an underlying 3D Gaussian \coarsegs via SDS loss~\cite{poole2022dreamfusion} against \synthesizer with the condition of \alignedimage, as shown in the left-bottom of~\cref{fig:pipeline}.

\paragraph{HairSynthesizer.}
We build \synthesizer based on \dataset to provide prior for unified 3D hair reconstruction.
With the same architecture as~\cite{liu2023zero}, our \synthesizer $S$ can generate the plausible novel view $I_n$ from random noises $N$ via a multi-step denoising process, conditioned on the pair of \alignedimage and an arbitrary relative camera pose $(R,T)$ . 
The operation can be defined as following:
\begin{align}
I_n = S(N | (I_a, R, T)).
\end{align}

\paragraph{3D Gaussian.}
To reconstruct 3D hair with the aid of \synthesizer, a 3D representation is needed.
Textured mesh~\cite{liu2023one}, NeRF~\cite{liu2023zero} and 3D Gaussian~\cite{tang2023dreamgaussian} are proved to be optional representations.
We choose 3D Gaussian~\cite{kerbl3Dgaussians} as the underlying 3D hair representation, because it is faster to be optimized than NeRF and has more flexible topology than mesh.
A 3D Gaussian model $\Theta$ consists of many Gaussian primitives, following~\cite{kerbl3Dgaussians}, we define the $i_{th}$ primitive $\Theta_i$ as $\Theta_i=\{x_i, s_i, q_i, \alpha_i, c_i\}$, where $x_i \in \mathbb{R}^3$, $s_i \in \mathbb{R}^3$, $q_i \in \mathbb{R}^4$, $\alpha_i \in \mathbb{R}$ and $c_i \in \mathbb{R}^3$ represents kernel center, scale, rotation, opacity and color, respectively. 
To reduce the parameters in the optimization, we directly optimize the RGB value of Gaussian primitives.

\paragraph{Optimization.}
We initialize the coarse Gaussian \coarsegs within a bounding box containing the upper-half of the template body. 
Following DreamGaussian\cite{tang2023dreamgaussian}, we optimize the corresponding \coarsegs of the hair and body in \alignedimage iteratively by minimizing a SDS loss $\mathcal{L}_{SDS}$ and a reference loss $\mathcal{L}_{ref}$.
The SDS loss is formulated as:
\begin{align}
    \nabla_{\Theta} \mathcal{L}_{SDS}=\mathbb{E}_{t,p,\epsilon}[(\epsilon_{S}(I_{\Theta^0}^{(R,T)};t,I_{a},(R,T))-\epsilon)\frac{\partial I_{\Theta^0}^{(R,T)}}{\partial \Theta^0}]
\end{align}
where $\epsilon_{S}$ is the predicted noise by \synthesizer $S$, and $(R,T)$ is the relative camera pose change from the reference view of \alignedimage. 
$I_{\Theta^0}^{(R,T)}$ represents the rendered RGB image of \coarsegs from the relative view $(R,T)$.
The reference loss is the L1 distance of RGB and transparency between $I_{a}$ and the rendered RGB image $I_{a}^{'}$ from the reference view by setting $R=T=0$, as well as their mask $M_{a}$ and $M_{a}^{'}$:
\begin{align}
    \mathcal{L}_{ref}=||I_{a}-I_{a}^{'}||_{1}+||M_{a}-M_{a}^{'}||_{1}.
\end{align}
After the optimization, we obtain \coarsegs representing the 3D hair and template body with consistent coarse shape and hair color.

\input{fig/synthesizer}

\subsection{View-wise Gaussian Refinement}
\paragraph{Motivation.}
The coarse Gaussian optimization can only recover the rough 3D hair because of the inherent view-inconsistency of \synthesizer.
Especially, the quality of rendering of \coarsegs is at a very low level.
The texture of $I_{\Theta^0}$ shows very noisy patterns (see~\cref{fig:synthesizer}).
Inspired by the diffusion-based image editing~\cite{meng2022sdedit}, we believe combining the advantages of obtained 3D hair \coarsegs and the prior of \synthesizer can alleviate the issue of texture.
\coarsegs provides consistent 3D information but bad texture, while \synthesizer can generate multi-view images with plausible texture but bad view-consistency.
Thus, to improve the texture quality of obtained 3D hair, we perform an view-wise Gaussian refinement using \coarsegs and \synthesizer jointly.

\paragraph{Refinement.}
The view-wise Gaussian refinement is conducted in optimization manner.
As shown in~\cref{fig:synthesizer}, at each step, we first render \coarsegs into an image $I_{\Theta^0}$ from a random view $(R,T)$.
Then we blend this image with random noises and obtain its refined version $I_{refine}$ through the multi-step denoising of \synthesizer conditioned on \alignedimage and the camera pose $(R,T)$:
\begin{align}
I_{refine} = S((\gamma \cdot I_{\Theta^0} + (1-\gamma) \cdot N) | (I_a, R,T)),
\end{align}
where $\gamma \in (0,1)$ is a blending coefficient enlarged during the optimization.
With a smaller $\gamma$, the generated $I_{refine}$ has better texture but worse alignment with $I_{\Theta^0}$, i.e. view-consistency ensured by \coarsegs.
On the contrast, with a big $\gamma$ approaching $1$, \synthesizer will lose the ability to refine the flawed texture of $I_{\Theta^0}$.
To optimize \coarsegs, we minimize the $L_1$ and perceptual loss between $I_{\Theta^0}$ and $I_{refine}$:
\begin{align}
\mathcal{L}_{refine}=||I_{\Theta^0}-I_{refine}||_{1}+\beta\cdot||\phi(I_{\Theta^0})-\phi(I_{refine})||_{1},
\end{align}
where $\phi(\cdot)$ denotes the operation of extracting VGG features~\cite{simonyan2014very}.
Please note that we describe the process with batch size 1 for convenience, but we render several views each step in practice for stable optimization and also add the supervision of $I_a$ for achieving better quality of frontal views.
After the view-wise Gaussian refinement, we obtain a refined 3D Gaussian \refinedgs. 
As shown in the zoom-in rendering comparison of~\cref{fig:synthesizer}, the texture quality of \refinedgs is improved a lot than \coarsegs, where noisy patterns are basically cleaned.

\input{fig/enhancer}

\subsection{Pixel-wise Gaussian Refinement}
\paragraph{Motivation.}
Even with the view-wise Gaussian refinement, the texture quality of \refinedgs is still unsatisfactory.
We believe the main reason is that \synthesizer focuses on novel-view generation.
During its training, the mis-alignment of overall hair appearance in view-wise influences much more than the correctness of pixel-wise texture details.
Based on such prior, the carefully distilled \refinedgs shows a blurry upper bound of hair texture.
To address this issue, we build a diffusion-based pixel-wise prior \deblurer and further conduct a pixel-wise Gaussian refinement.

\paragraph{HairEnhancer.}
We build \deblurer based on \dataset to provide prior for hair texture enhancement.
Conditioning on a blurry hair image $I_{blur}$, \deblurer can generate the detail-enhanced image $I_{enhance}$ with strand-like textures from random noises $N$ via a multi-step denoising process. 
The operation can be defined as $I_{enhance} = E(N | I_{blur})$.

\paragraph{Refinement.}
As shown in~\cref{fig:enhancer}, similar to view-wise refinement, we perform the pixel-wise Gaussian refinement through optimizing the \refinedgs via minimizing the loss:
\begin{small}
\begin{align}
\mathcal{L}_{enhance}=||I_{\Theta^1}-E(N | I_{\Theta^1})||_{1}+\beta\cdot||\phi(I_{\Theta^1})-\phi(E(N | I_{\Theta^1}))||_{1}.
\end{align}
\end{small}
Comparing to \refinedgs, \enhancedgs shows fine hair texture with strand-like details.
The obtained final 3D Gaussian \enhancedgs can achieve fast and high-quality multi-view rendering via Gaussian splatting~\cite{kerbl3Dgaussians}.
Furthermore, in \enhancedgs, any 3D hairstyles are aligned with the same template body. In other words, our reconstructed 3d hairstyles are normalized and canonical, which may further enable the creation of a large-scale 3D hair dataset aligned with real images.

%% file: fig/synthesizer.tex
\begin{figure}
\centering
\includegraphics[width=1.0\linewidth]{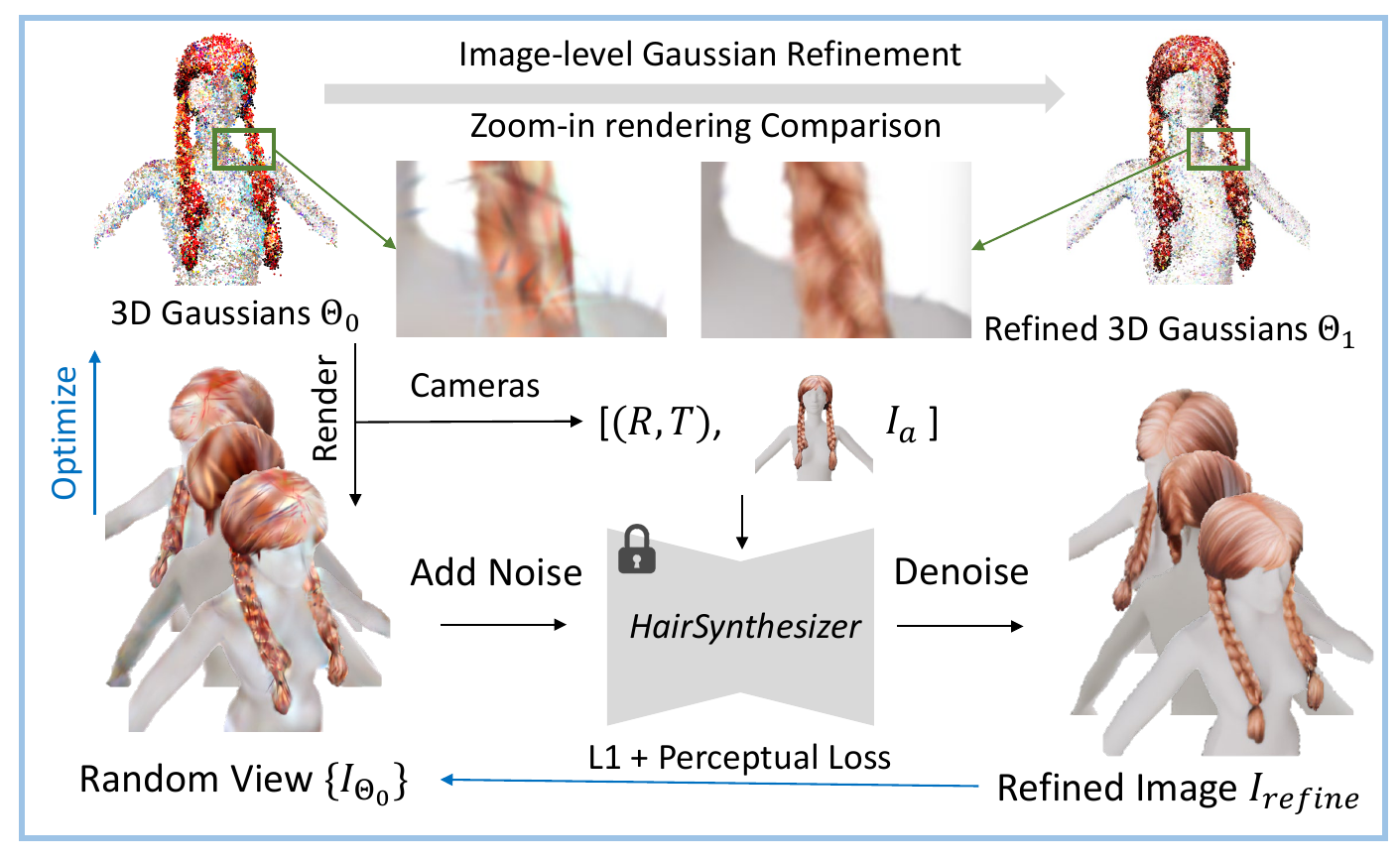}
\caption{With the aid of \synthesizer, we perform image-wise Gaussian refinement on optimized coarse 3D Gaussian. The refined 3D Gaussian will be obtained with better texture.}
\label{fig:synthesizer}
\end{figure}

%% file: fig/enhancer.tex
\begin{figure}
\centering
\includegraphics[width=1.0\linewidth]{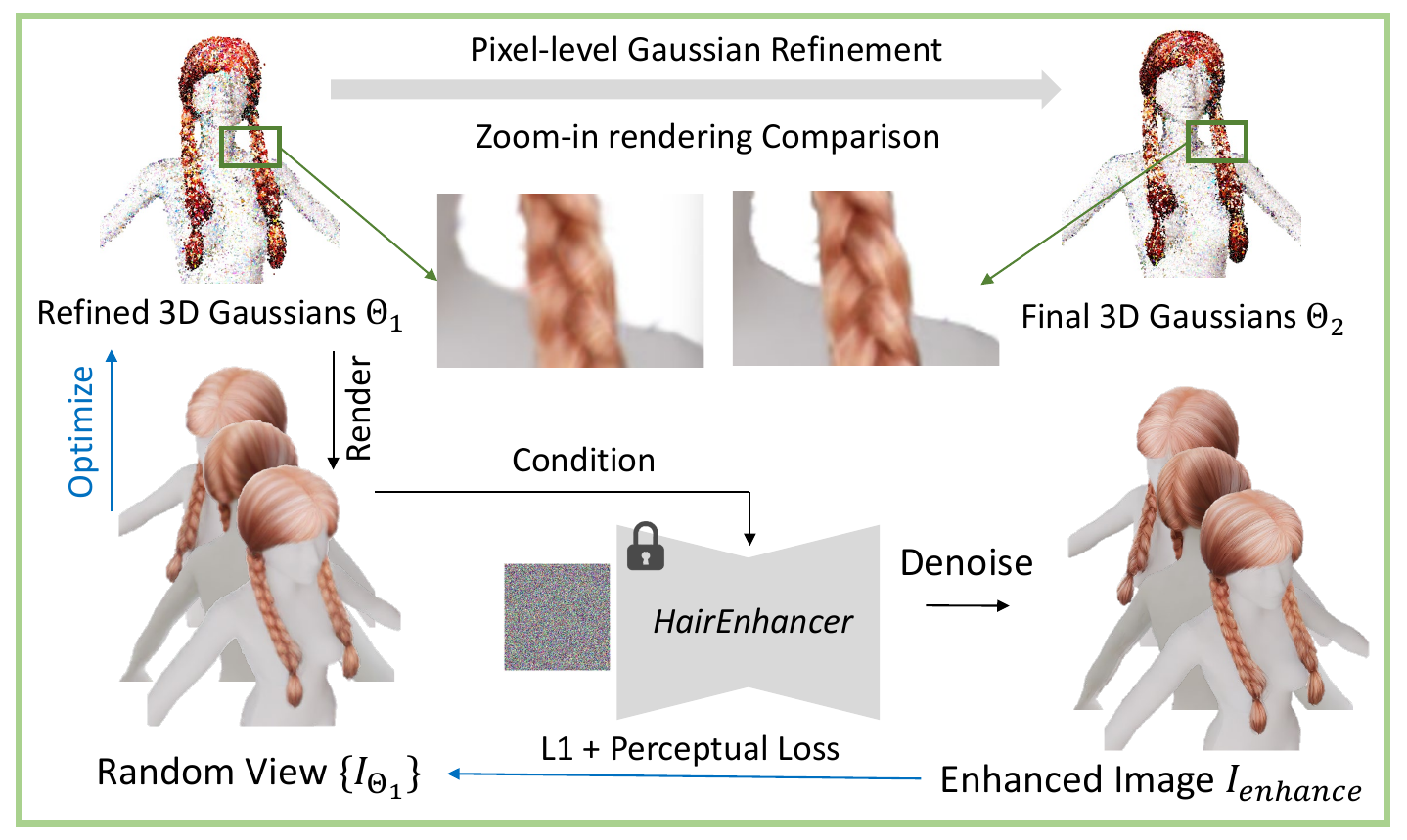}
\caption{With the aid of \deblurer, we perform pixel-wise Gaussian refinement on optimized coarse 3D Gaussian. The enhanced 3D Gaussian will be obtained with better texture.}
\label{fig:enhancer}
\end{figure}

%% file: sec/experiments.tex
\input{fig/compare_svr}

\section{Experiments}
\label{sec:experiments}

\subsection{Data Split and Evaluation Metrics}
We split the 2,396 3D hair samples in \dataset into training and test partitions with a ratio of about 9:1.
Thus, the training set for \synthesizer and \deblurer contains 2,155 3D hairstyles including 73,862 different textured hair meshes. 
The remaining 241 3D hairstyles are used to evaluate the proposed method.

To evaluate our method quantitatively, we calculate the $L_1$ error, $PSNR$ and $Perceptual$ error~\cite{johnson2016perceptual} within the region of ground-truth hair mask for 241 hairstyles in the synthetic testing set, where three texture maps per hairstyle are randomly chosen (723 textured meshes) and 20 views per mesh are randomly selected to render (14,460 images). 
For the perceptual error, we compute the sum of $L_1$ distance of the features after layer 4, 9, 16 and 23 of VGG16~\cite{simonyan2014very} between the rendered images of final 3D Gaussian \enhancedgs and the ground-truth images.

\subsection{Comparisons}
\paragraph{Comparisons with single-view 3D hair reconstruction.}
We compare our method with single-view hair reconstruction methods~\cite{zheng2023hairstep},~\cite{hu2017avatar} and~\cite{sun2021single} on braided styles in~\cref{fig:compre_failure}.
~\cite{zheng2023hairstep} fails on the hairstyle of braids and buns (see~\cref{fig:compre_failure} (a)), because of lacking corresponding 3D strand-level braided data in its underlying hair prior, which hinders its generalization capability to diverse real-world hairstyles.
As shown in~\cref{fig:compre_failure} (b), the method of~\cite{hu2017avatar} can produce textured 3D hair strips.
However, it fails to recover the structure of braids and the texture of their results are lack of realism. 
We also compare our strategy with the braid-specific method~\cite{sun2021single}.
Our method can successfully reconstruct the double-braid 3D hairstyle faithfully, while~\cite{sun2021single} fails to recover the invisible parts of the hair,which is emphasized by the red box in~\cref{fig:compre_failure} (c).
Also, the texture quality of our result is much better than~\cite{sun2021single}.
Please note that, \cite{sun2021single} is not an automatic approach and our alignment method based on facial landmarks cannot work for the input from back-view.
In this example, we use a same semi-automatic method to do the alignment as~\cite{sun2021single}.  

\input{fig/compare_real}

\paragraph{Comparisons with diffusion-based 3D reconstruction.}
To further demonstrate the merits of our system on single-view hair modeling, we compare our strategy with most recent generative Image-to-3D works DreamGaussian~\cite{tang2023dreamgaussian}, One-2-3-45~\cite{liu2023one} and One-2-3-45++~\cite{liu2023one1}, given front-view condition on in-the-wild real images. 
For the visual comparison in ~\cref{fig:compare_real}, the results of previous methods are full of various artifacts, such as irrational hairstyle, undesired shape and blurry texture. 
On the contrast, our method achieves faithful 3D hair reconstruction.
Moreover, although our priors \synthesizer and \deblurer are trained on synthetic data, our system has fine performance on in-the-wild images involving highly diverse hairstyles with a non-negligible domain gap, as shown in~\cref{fig:compare_real} and~\cref{fig:more_results}. 
To demonstrate the superior of our method, we conduct a user study on 10 randomly selected hairstyles, with total 32 volunteers involved.
In the questionnaires, they are asked to evaluate the rationality and the texture quality of randomly sorted reconstructed 3D hairstyles of different approaches, by giving integer scores from 0 to 10.
The final average scores of all methods are listed in~\cref{tab:user_study}, where ours is ranked the best, supporting the analysis of the visual results.
More results and visualization of 3D Gaussians on real-world images can be found in the supplementary video. 

\input{fig/user_study}

\input{fig/ablation}

\subsection{Ablation Study}
\paragraph{Ablation on Gaussian refinement.}
To better analyze the design of different modules of our pipeline, we compare the results of \synthesizer and the rendering results of \coarsegs, \refinedgs and \enhancedgs from real inputs in~\cref{fig:ablation}, qualitatively.
From the comparisons between coarse Gaussian \coarsegs (b) and refined Gaussian \refinedgs (c) in~\cref{fig:ablation}, we can easily see the effectiveness of view-wise Gaussian refinement.
Furthermore, after the process of pixel-wise Gaussian refinement, the texture of \enhancedgs (d) will be clearly improved, comparing with \refinedgs (c).
Please check the zoom-in details in (e).
We also show the results of our diffusion prior \synthesizer in\cref{fig:ablation}(a), where view-inconsistency occurs.
For each example, images from two very near view angles are given in the first and the second row, respectively.
In the red boxes of the upper example, inconsistent hair geometries are presented, the first view shows ridge-like shape while the second view shows contrary valley-like shape.
In terms of the lower example, the comparisons between the yellow boxes and red boxes present inconsistent texture and shape, respectively.
With the help of the underlying 3D Gaussian, our results performs well in view-consistency.
\cref{tab:ablation} provides quantitative supports by evaluating on the test subset of \dataset.

\input{fig/ablation_T}

\input{fig/ablation_tab}
\input{fig/ablation_T_tab}

\paragraph{Ablation on enlarging $\gamma$.}
In the view-wise Gaussian refinement, we gradually enlarging the weight $\gamma$ of the images rendered from the being optimized coarse Gaussian \coarsegs, and add less noise into the input of diffusion prior \synthesizer.
As shown in~\cref{tab:ablation_T}, starting from 0.5, we add 0.15 to $\gamma$ every 200 steps. 
From the comparisons in~\cref{fig:ablation_T}, the noising pattern of the texture will gradually removed using this strategy.
Comparisons in~\cref{tab:ablation} also verifies our observation.

\input{fig/more_results}

\subsection{Applications}

\paragraph{Boosting single-view strand reconstruction.}
Our strategy has the ability to generate consistent and high-quality multi-view hair images from a single input.
As a result, with the help of our method, the single-view 3D strand reconstruction can be converted to maturer multi-view reconstruction~\cite{sklyarova2023neural}.
We feed 180 images rendered by our enhanced 3D Gaussian \enhancedgs to multi-view reconstruction methods~\cite{sklyarova2023neural,wu2024monohair} and finally obtain reconstructed 3D hair strands which is more reasonable than the results of the state-of-the-at approach~\cite{zheng2023hairstep} as shown in~\cref{fig:strand}.

\input{fig/strand}

\input{fig/generalize_side}

\paragraph{Generalization to side-view inputs.}
Our method is based on good facial landmark detection which is quite simple for frontal view images but difficult for side views.
As shown in~\cref{fig:generalize_side}, to make our method work for side-view input, we first replace the body of the input image with grey mask, then we generate the reference view, i.e. $R=T=\textbf{0}$, using \synthesizer conditioned on this masked hair image.
Finally, we input the synthesized aligned hair image of input side-view image to our system and reconstruct 3D hair.
Rendered images of outputted 3D Gaussian from two random views are given in~\cref{fig:generalize_side} (d-e).

\paragraph{Hairstyle customization for 3D avatars.}
As shown in~\cref{fig:avatar}, given a 3D avatar, we can customize its hairstyle to match the styles of reference images. We first reconstruct the target 3D hair from the single image via our method, which is aligned with our 3D body template. This hairstyle can then be applied to the given 3D avatar, which is also aligned with our template.

\input{fig/avatar}

%% file: fig/compare_svr.tex
\begin{figure}
\centering
\includegraphics[width=1.0\linewidth]{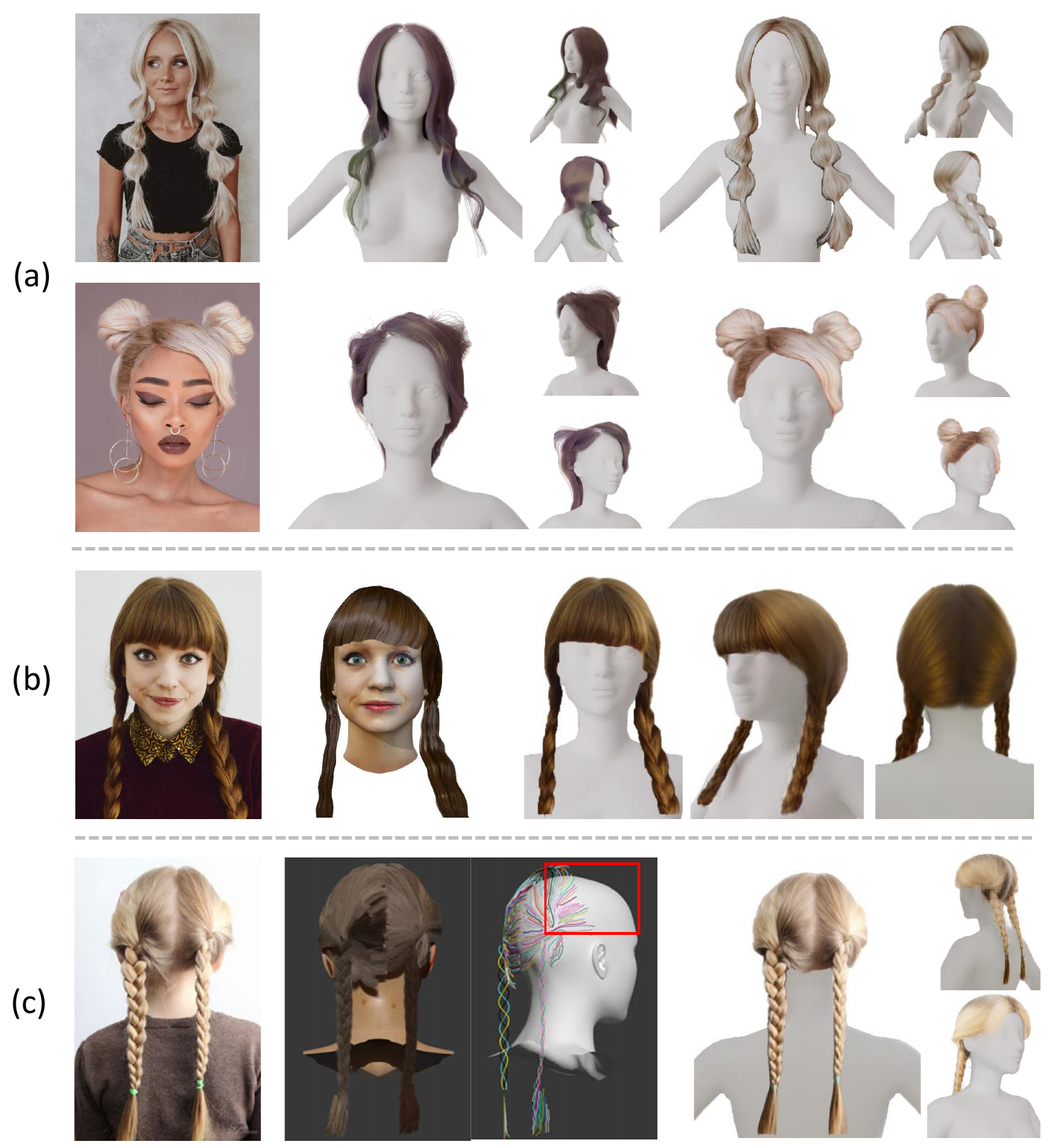}
\caption{Comparison with (a)~\cite{zheng2023hairstep}, (b)~\cite{hu2017avatar}, and (c)~\cite{sun2021single} on braided hairstyles. From left to right: input images, results of previous methods and ours.}
\label{fig:compre_failure}
\end{figure}

%% file: fig/compare_real.tex
\begin{figure}
\centering
\includegraphics[width=1.0\linewidth]{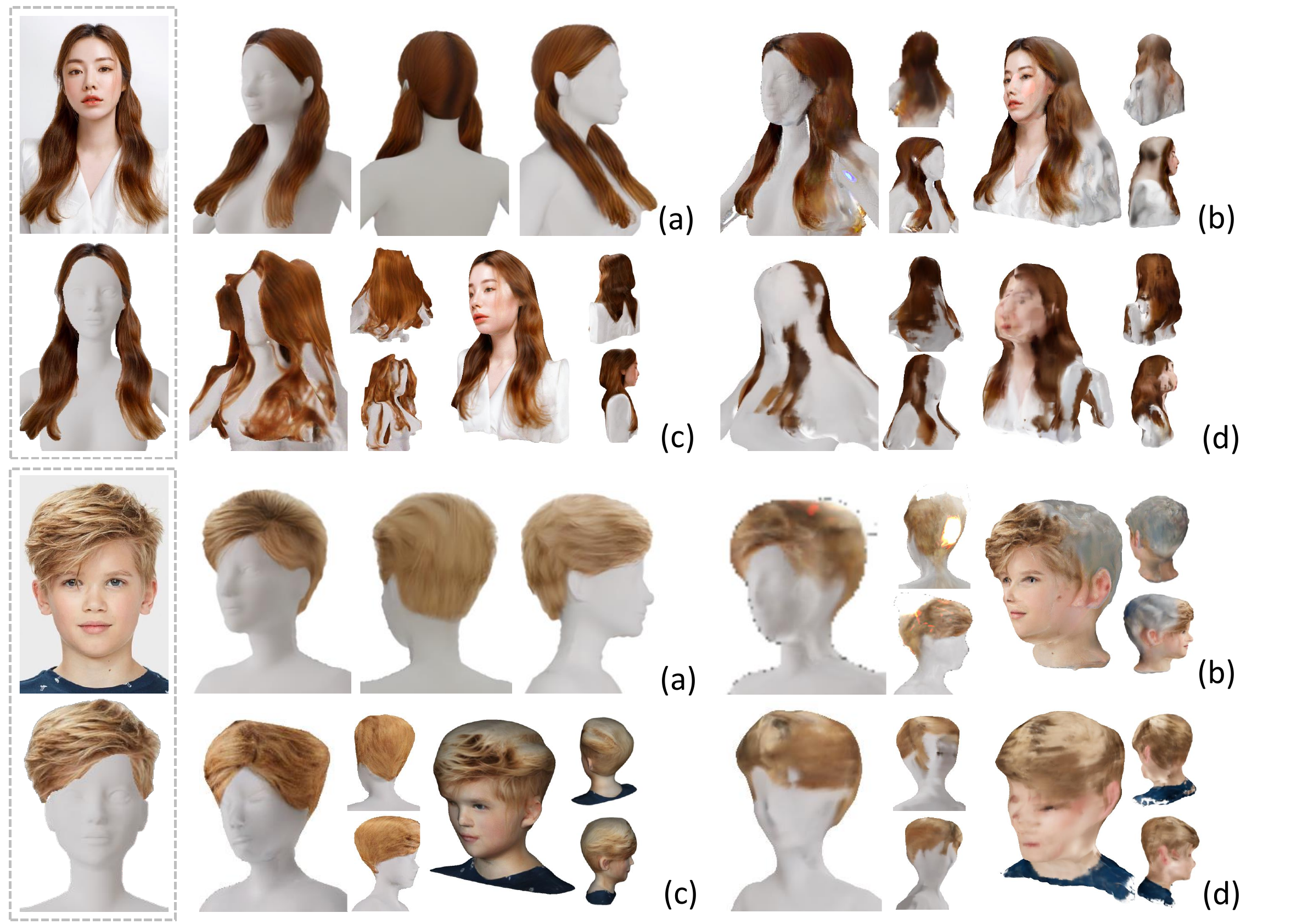}
\caption{Comparisons with diffusion-based 3D reconstruction. In the left dashed box, there are reference image and the aligned image of given hair and our template body. We compare the reconstructed 3D hair of (a) our method, (b) DreamGaussian\cite{tang2023dreamgaussian}, (c) One-2-3-45++\cite{liu2023one1} and (d) One-2-3-45\cite{liu2023one}. For previous methods, we provide results with inputs of both aligned images and real portraits.}
\label{fig:compare_real}
\end{figure}

%% file: fig/user_study.tex
\begin{table}
\centering
  \begin{tabular}{l|c|c|c|c}
   \hline
        {Methods}& Ours & \emph{DreamGaussian} & \emph{One-2-3-45++} & \emph{One-2-3-45} \\
    \hline
    {$Rat.$ $\uparrow$} & \textbf{9.08} & 4.52 & 6.33 & 2.36 \\
    {$Tex.$ $\uparrow$} & \textbf{8.96} & 4.25 & 6.30 & 2.09 \\
   \hline
  \end{tabular}
  \caption{Statistics of user study on reconstructed 3D hair by different methods. $Rat.$ and $Tex.$ stand for the rationality and texture quality.}
 \label{tab:user_study}
\end{table}

%% file: fig/ablation.tex
\begin{figure}
\centering
\includegraphics[width=1.0\linewidth]{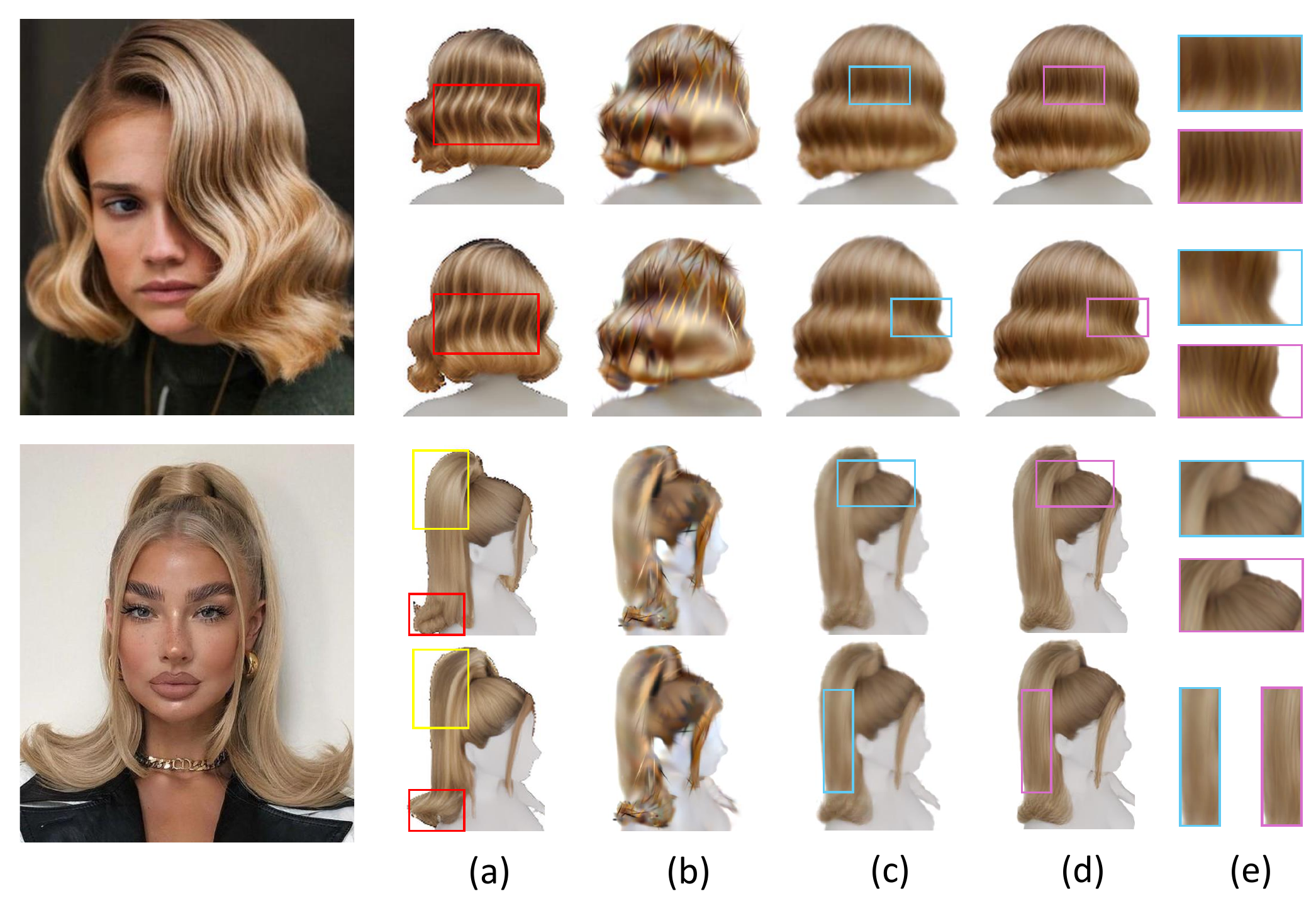}
\caption{Ablation study. (a) is the output of \synthesizer. (b-d) present the rendering results of \coarsegs, \refinedgs and \enhancedgs, respectively. (e) is the zoom-in comparison between \refinedgs and \enhancedgs. The first row and the second row show the results from two close-up viewpoints.}
\label{fig:ablation}
\end{figure}

%% file: fig/ablation_T.tex
\begin{figure}
\centering
\includegraphics[width=1.0\linewidth]{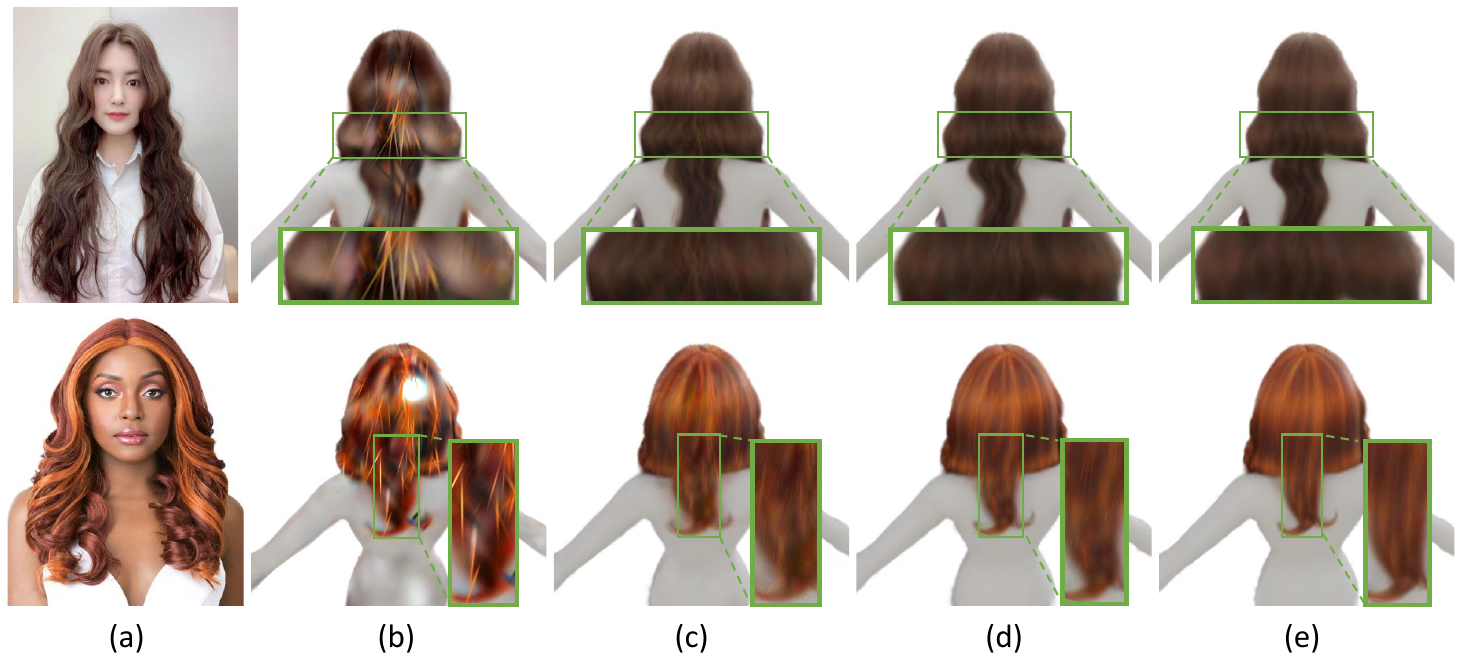}
\caption{Ablation study on enlarging $\gamma$ during image-wise Gaussian refinement. (a) Input images. (b) Rendering images of coarse Gaussian \coarsegs. (c-e) Rendering images of refined Gaussian \refinedgs after 200 steps ($\gamma=0.5$), 400 steps ($\gamma=0.65$) and 600 steps ($\gamma=0.8$), respectively. }
\label{fig:ablation_T}
\end{figure}

%% file: fig/ablation_tab.tex
\begin{table}
\centering
  \begin{tabular}{l|c c c c}
   \hline
   Method & \synthesizer & \coarsegs & \refinedgs & \enhancedgs (Full) \\
        \hline
        $L_1$ $\downarrow$ & 0.01565 & 0.02377 & 0.01249 & \textbf{0.01236}  \\
        $Perc.$ $\downarrow$ & 3.262 & 4.427 & 2.403 & \textbf{2.352} \\
        $PSNR$ $\uparrow$ & 26.07 & 24.92 & 28.99 & \textbf{29.60}  \\
   \hline
  \end{tabular}
  \caption{Ablation on Gaussian refinement. $Perc.$ denotes perceptual errors.}
 \label{tab:ablation}
\end{table} 

%% file: fig/ablation_T_tab.tex
\begin{table}
 \centering
  \begin{tabular}{l|c c c c}
   \hline
   \multirow{2}{*}{Method} & \coarsegs & $step\rightarrow200$  & $step\rightarrow400$ & $step\rightarrow600$ \\
                        &  & $\gamma=0.5$ & $\gamma=0.65$ & $\gamma=0.8$ \\ \hline
        $L_1$ $\downarrow$ & 0.02377 & 0.01298 & 0.01252 & \textbf{0.01249}  \\
        $Perc.$ $\downarrow$ & 4.427 & 2.559 & 2.461 & \textbf{2.403} \\
        $PSNR$ $\uparrow$ & 24.92 & 28.79 & 28.90 & \textbf{28.99}  \\
   \hline
  \end{tabular}
  \caption{Comparisons on enlarging $\gamma$ during view-wise Gaussian refinement. $Perc.$ denotes perceptual errors.}
 \label{tab:ablation_T}
\end{table}

%% file: fig/more_results.tex
\begin{figure*}
\centering
\includegraphics[width=1.0\textwidth]{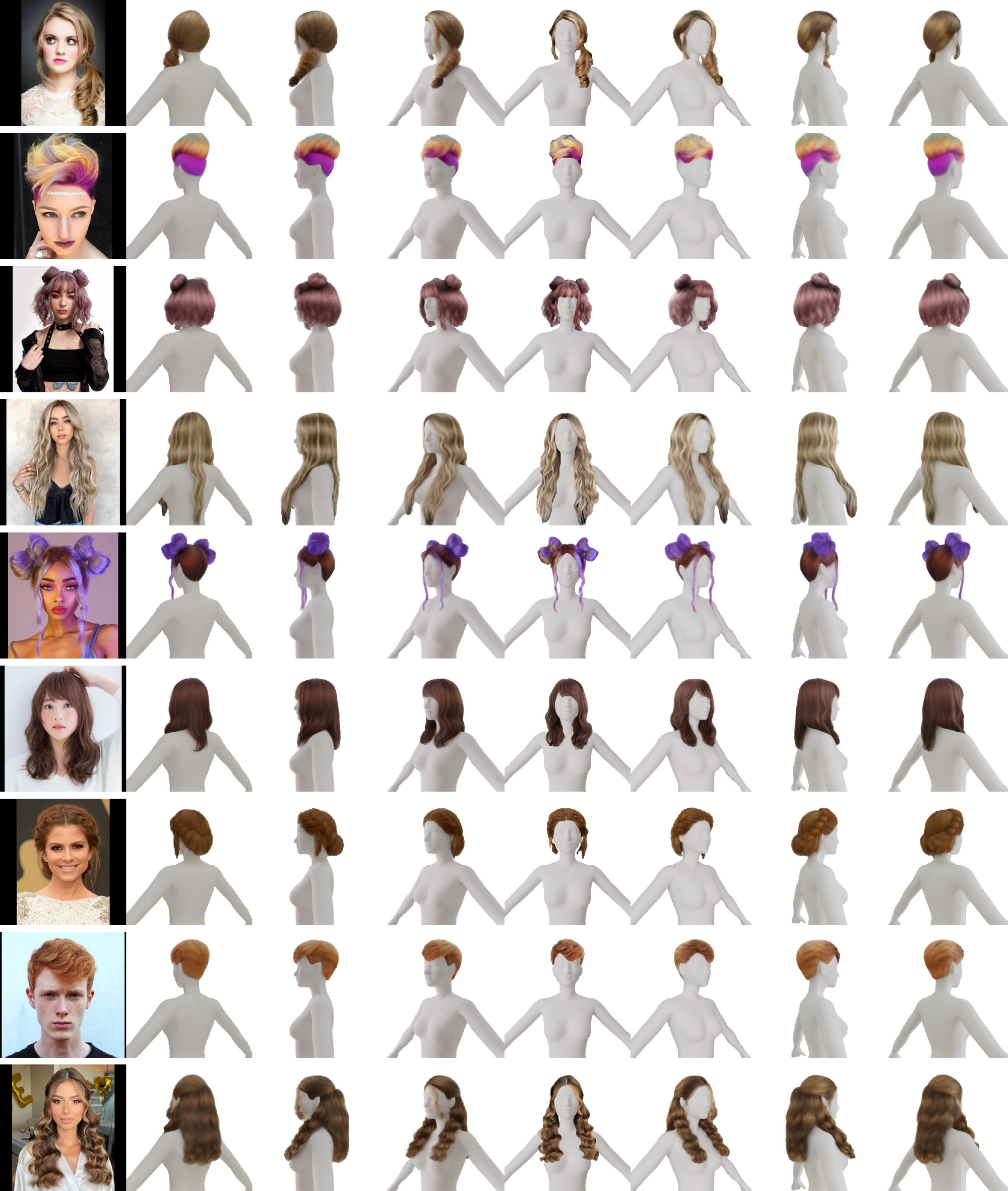}
\caption{More results on in-the-wild images. We render our reconstructed 3D hair from 7 views for illustration.}
\label{fig:more_results}
\end{figure*}

%% file: fig/strand.tex
\begin{figure}
\centering
\includegraphics[width=\linewidth]{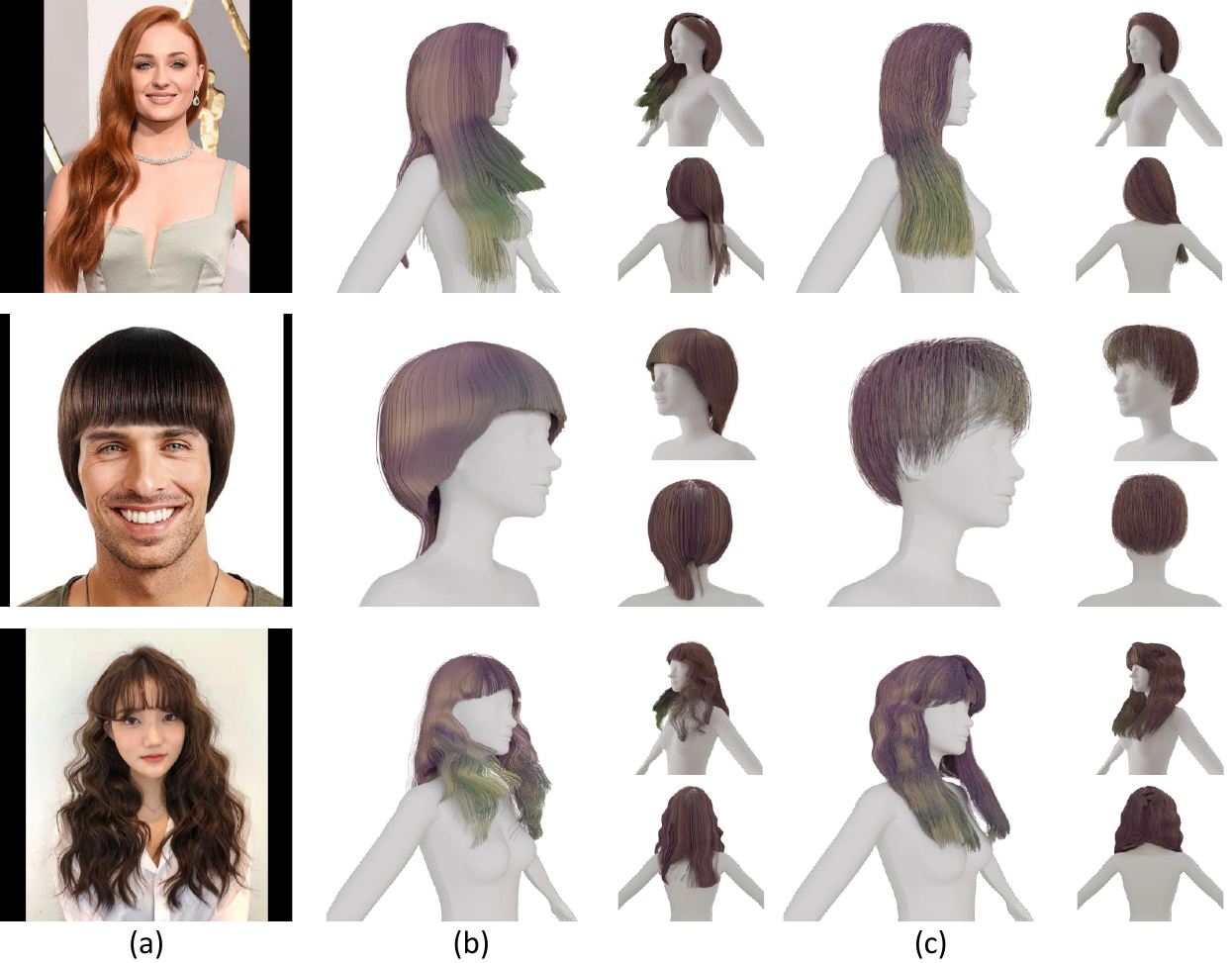}
\caption{Application for single-view 3D strand reconstruction. (a) Reference images. (b) Results of~\cite{zheng2023hairstep}. (c) Our results reconstructed by combining our method and multi-view reconstruction methods (NeuralHairCut~\cite{sklyarova2023neural} for the first two rows and MonoHair~\cite{wu2024monohair} for the last).}
\label{fig:strand}
\end{figure}

%% file: fig/generalize_side.tex
\begin{figure}
\centering
\includegraphics[width=\linewidth]{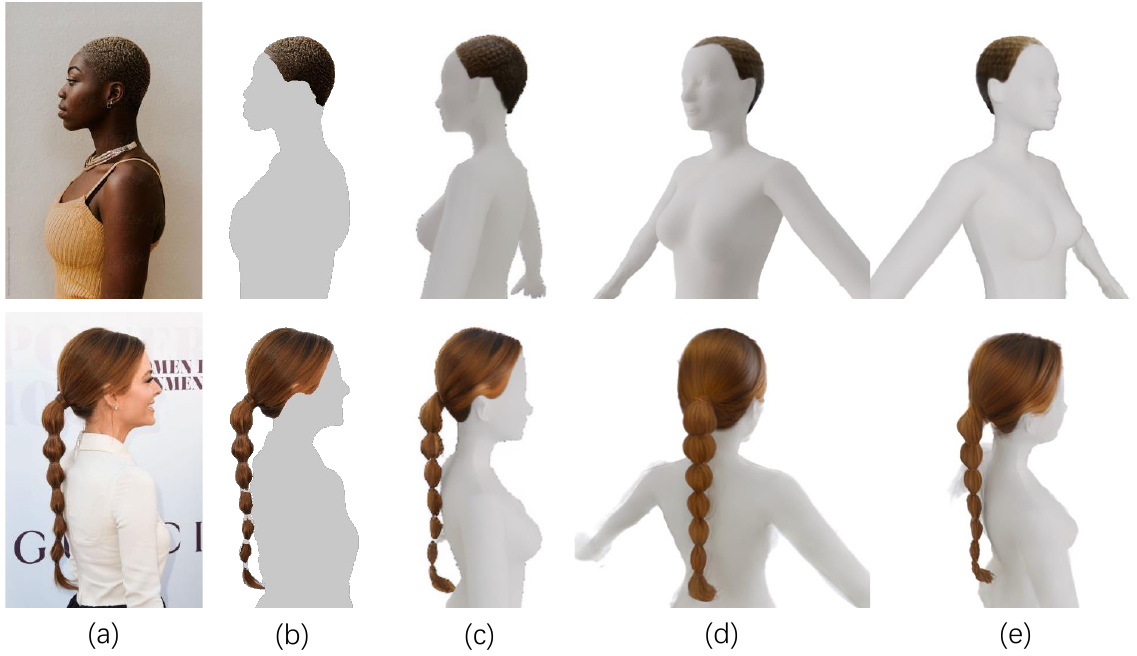}
\caption{Reconstructed results given side-view inputs. (a) Input images. (b) Masked hair with grey body. (c) The output image of reference view of \synthesizer conditioned on (b). (d-e) Output images of our method taking (c) as input.}
\label{fig:generalize_side}
\end{figure}

%% file: fig/avatar.tex
\begin{figure}
\centering
\includegraphics[width=1.0\linewidth]{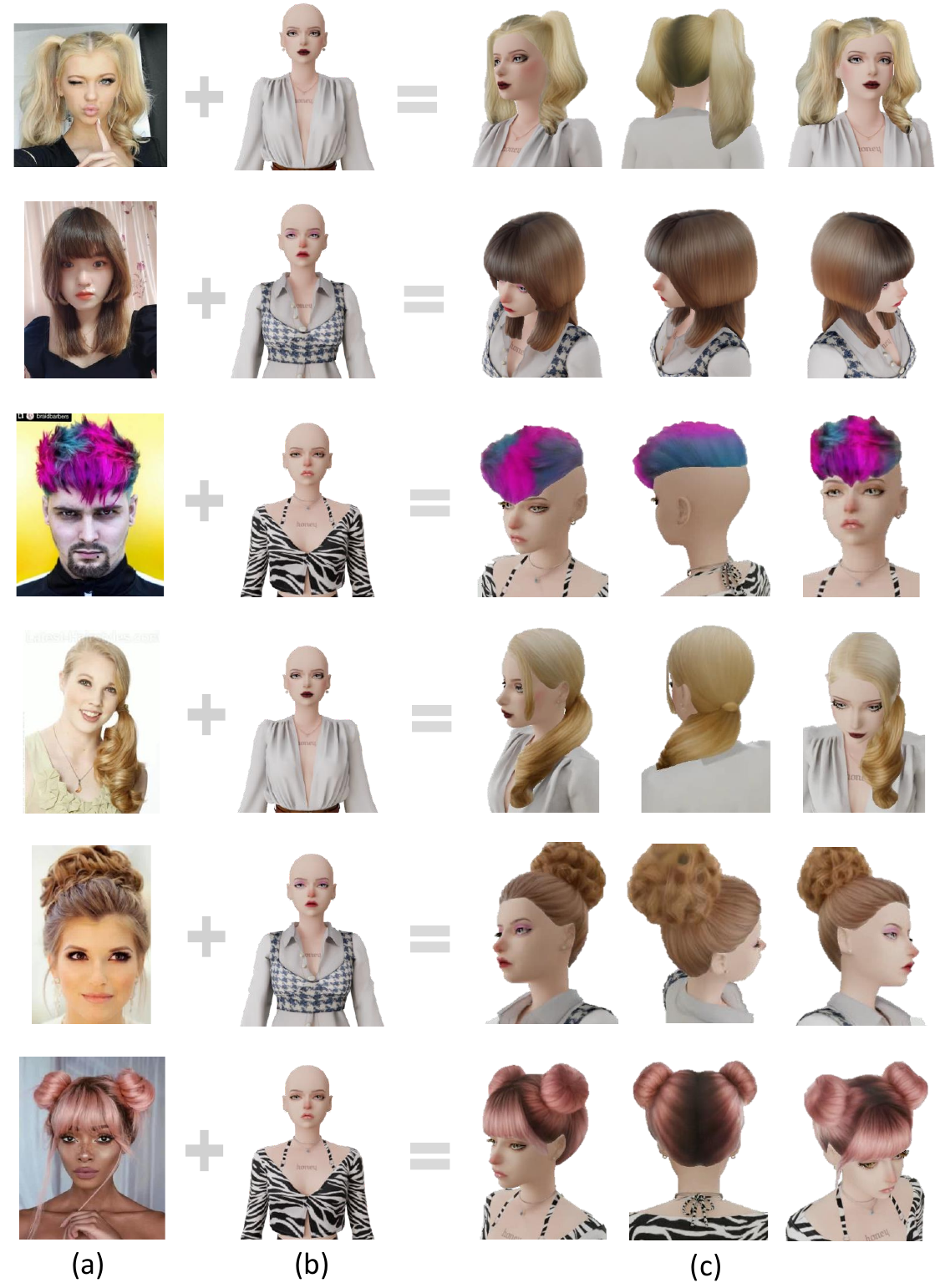}
\caption{Application for changing hair of 3D avatars. (a) Reference images. (b) 3D avatars. (c) Rendered images of (b) with the hair in (a).}
\label{fig:avatar}
\end{figure}

%% file: sec/conclusion.tex
\section{Conclusion}
\label{sec:conclusion}
In this work, we propose a novel strategy towards unified 3D hair reconstruction from single-view images. 
Our method adopts an optimization-based 2D lifting approach to reconstructed hair using 3D Gaussian, enhanced by two novel hair-specific diffusion priors, \synthesizer and \deblurer.
To train these two priors, we collect a large-scale 3D hair dataset featuring a variety of various braided and un-braided hairstyles and render hair images from multiple views.
With two carefully designed modules, i.e., view-wise and pixel-wise Gaussian refinement, our method can reconstruct faithful 3D hair with consistent and high-quality textures.
Our extensive experiments show the capability of reconstructing diverse 3D hairstyles in a unified pipeline, and highlight the superiority of our approach compared to existing methods.

\input{fig/limitation}

\paragraph{Limitations.}
Although our approach yields realistic results on a wide range of input images, it may fail in certain cases when tested on challenging portraits. 
For instance, as shown in~\cref{fig:limitation}, our method relies on successful intermediate stages such as (a) hair-head alignment and (b) hair mask segmentation.
Inaccuracies in these processes can adversely affect the final results. 
Also, our method may struggle with complex real-world lighting, which can impact hair texture (see the top of the hair in~\cref{fig:limitation}(c)).  
Hair accessories may cause inconsistencies between different views (\cref{fig:limitation}(d)).
Our method does not yet perform well on wavy, curly, and coily hair, which we hope to improve in the future.
Additionally, although our method shows generalization capability on real portraits, it still needs improvement to achieve high fidelity. We believe this is mainly caused by the usage of limited synthetic data. A promising direction is leveraging priors from massive high-quality 2D real images for 3D hair reconstruction.

\begin{acks}
This work was supported in part by the Basic Research Project No. HZQB-KCZYZ-2021067 of Hetao Shenzhen-HK S\&T Cooperation Zone, Guangdong Provincial Outstanding Youth Fund(No. 2023B1515020055), the National Key R\&D Program of China with grant No. 2018YFB1800800, by Shenzhen Outstanding Talents Training Fund 202002, by Guangdong Research Projects No. 2017ZT07X152 and No. 2019CX01X104,by Key Area R\&D Program of Guangdong Province (Grant No. 2018B030338001), by the Guangdong Provincial Key Laboratory of Future Networks of Intelligence (Grant No. 2022B1212010001), and by Shenzhen Key Laboratory of Big Data and Artificial Intelligence (Grant No. ZDSYS201707251409055).  This work is also partly supported by NSFC-61931024, and Shenzhen Science and Technology Program No. JCYJ20220530143604010.
\end{acks}


%% file: fig/limitation.tex
\begin{figure}
\centering
\includegraphics[width=\linewidth]{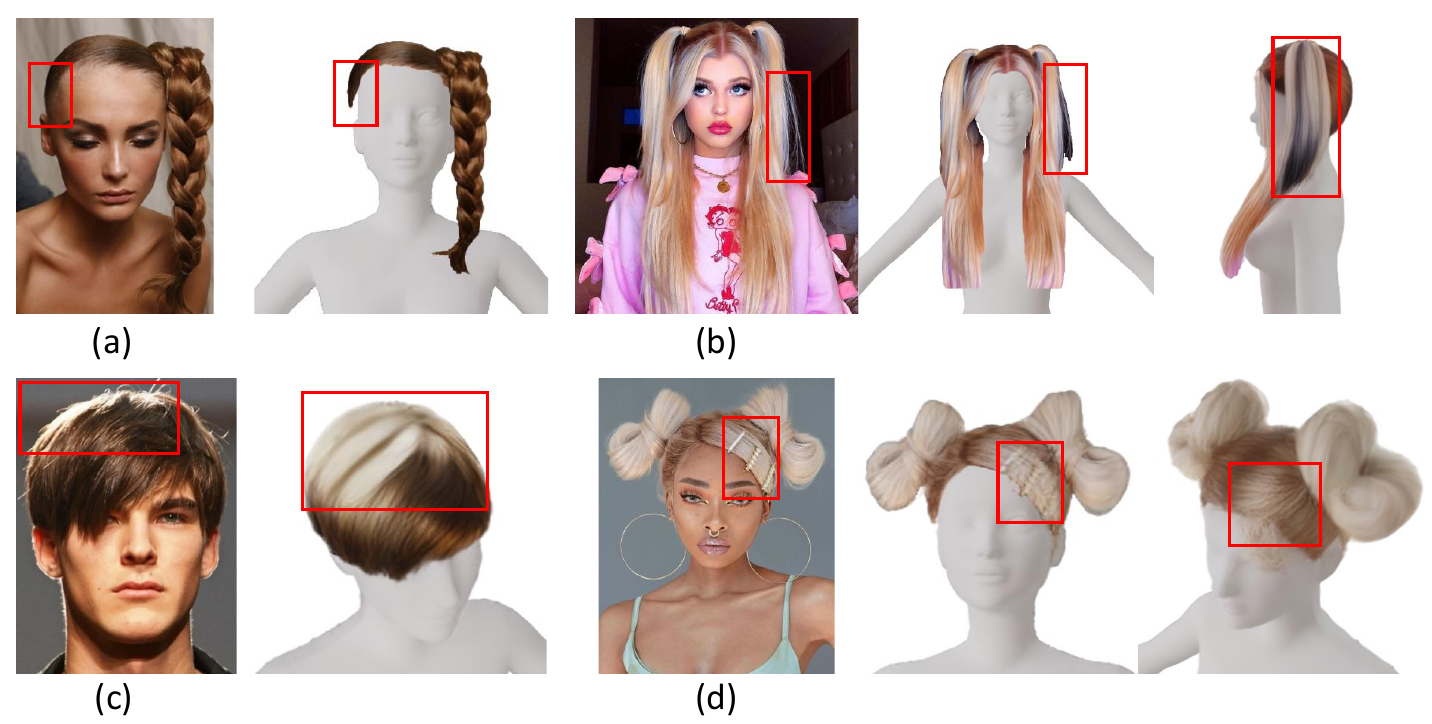}
\caption{Limitations on (a) hair-head mis-alignment, (b) inaccurate mask, (c) complex real-world lighting condition, and (d) hair accessories.}
\label{fig:limitation}
\end{figure}

%% file: sec/appendix.tex
\section{Implementation Details}
We describe the training details for diffusion priors \synthesizer and \deblurer, as well as the settings of Gaussian optimization for 3D hair reconstruction in this section.

\paragraph{{Diffusion Training.}} We sample 5 random views (including the frontal) from the upper semi-sphere per texture map of each 3D hair model with camera focal length=50mm and radius=1.05, resulting in 413,410 rendered images with the size of $512\times512$, where 369,310 images belongs to 2155 3D hair models in the training split of \dataset are for training. We set training step to 20, 000, batch size to 16. Using the same network architecture and default loss weight as~\cite{liu2023zero}, the training of \synthesizer requires about 12 hours on 8 NVIDIA RTX3090Ti.
\deblurer uses the similar network as \synthesizer, but removing the clip for camera encoding, because it only outputs the clearer images of conditioned reference views.
To guarantee the coverage of the full views, We use 180 dense views of each 3D hair model with a randomly selected texture map to train the \deblurer, including 387,900 images. 
Training setting and time of \deblurer are the same as \synthesizer.

\paragraph{{Gaussian Optimization.}} Coarse Gaussian optimization takes 5,000 randomly sampled points in the box of $\pm0.3$.
We follow~\cite{tang2023dreamgaussian} to densify the Gaussian every 100 iterations with the gradient of 0.01.
The pruning will be conducted after each operation of densification.
We set the learning rate of the position, color, opacity, scale and rotation of 3D Gaussian to 0.001, 0.01, 0.05, 0.005 and 0.005, respectively.
The learning rate will be reduced to $2\times10^{-5}$ gradually during the optimization.
The optimization of coarse 3D Gaussian runs 1,000 iteration with batch size 4 on single NVIDIA RTX3090Ti, which takes about 2 minutes.
View-aware and pixel-aware Gaussian refinement takes 600 and 1,000 iterations, costing about 40 minutes in total. 
We densify the Gaussian every 200 iterations with the gradient of 0.0002 during the Gaussian refinement.
We use the same learning rate as the coarse Gaussian optimization.
The camera setting for render 3D Gaussian is the same as our dataset preparation.


\section{\dataset Dataset}
We compare our \dataset with previous 3D hair dataset in~\cref{tab:compare_dataset}, including USC-HairSalon~\cite{hu2015single}, following datasets from recent hair modeling works~\cite{wu2022neuralhdhair, zhou2023groomgen, shen2023ct2hair} and commercial website DataGen \footnote{\url{{https://datagen.tech/}}} on human-centric data synthesis. So far, the largest open-source 3D hair dataset USC-HairSalon contains 343 hairstyles. It can only cover basic categories, such as long or short, straight or wavy, but fail to include complex types like extremely curly hair or braided hair. Following works on hair modeling also suffer from data shortage. To tackle on this, we construct our large-scale highly diverse synthetic hair
dataset \dataset of calibrated multi-view images. The scale of both geometries and textures are much larger than previous dataset.

We argue that although we do not provide strand-level geometry, the rich-textured and shape-diverse \dataset in multi-view representation is capable to support the learning of diffusion prior of 3D hairstyle. We show more hair samples in our \dataset with various styles (\cref{fig:dataset_shape}) , texture (\cref{fig:dataset_tex}) and rendering angles (\cref{fig:dataset_view}). 

\input{table/compare_dataset}
\input{fig/dataset_gallery}


\section{Ablation on Perceptual Loss}
We compare the rendering results with (w/) and without (w/o) using perceptual loss~\cite{johnson2016perceptual} (features of layer 4, 9, 16 and 23 of VGG16~\cite{simonyan2014very}).
As shown in~\cref{fig:ablation_vgg}, the perceptual loss helps a lot to obtain fine textures of \refinedgs and \enhancedgs in the process of view-wise Gaussian refinement and pixel-wise Gaussian refinement.
\cref{tab:ablation_vgg_tab} displays the quantitative comparisons supports this analysis.

\input{fig/ablation_vgg}
\input{table/ablation_vgg}


\section{Comparisons with DreamGaussian}
We compare our method with the related method DreamGaussian~\cite{tang2023dreamgaussian} in \cref{fig:compare_dreamgs} and \cref{tab:compare_dreamgs_tab}. 
From the comparisons, we find the results of DreamGaussian has lots of artifacts as shown in \cref{fig:compare_dreamgs} (a), mainly caused by the inappropriate general prior of Zero-1-to-3. With the help of our hair-specified prior \synthesizer, the results of DreamGaussian turn better (\cref{fig:compare_dreamgs} (b)), but still unsatisfactory.
We believe that this is because of the inflexibility of 3D mesh representation used in the refinement stage of~\cite{tang2023dreamgaussian}.
Also, the extraction of the coarse mesh from 3D Gaussian has lost quite amount of geometric details. 
The reconstructed 3D hair of our method is shown in \cref{fig:compare_dreamgs} (c).
It indicates that 3D Gaussian brings large flexibility in both shape and texture optimization, which makes the reconstruction of 3D hair a higher quality. 

\input{fig/compare_dreamgs}
\input{table/compare_dreamgs}


\section{More In-the-wild Results}
\input{fig/more_results1}
\input{fig/more_results2}

In this section, we provide more in-the-wild results of our approach in~\cref{fig:more_results1} and~\cref{fig:more_results2}, which demonstrates that our method has good generalization ability on real portraits, although using priors learned from synthetic data.

Please check our supplementary video for more detailed multi-view exhibitions.

%% file: table/compare_dataset.tex
\begin{table*}[htbp]
 \begin{center}
  \begin{tabular}{l|c|c|c|c}
   \hline
   Dataset & Open-source & Styles & Number of 3D hair & Textures\\
        \hline
        \emph{USC-HairSalon}~\cite{hu2015single} & \Checkmark & un-braided & 343 & \XSolidBrush\\
        \hline
        \emph{NeuralHDHair}~\cite{wu2022neuralhdhair} & \XSolidBrush & un-braided & 653 & \XSolidBrush\\
        \hline
        ~\cite{zhou2023groomgen} & \XSolidBrush & un-braided & 35 & \XSolidBrush\\
        \hline
        ~\cite{shen2023ct2hair} & \Checkmark & un-braided & 10 & \XSolidBrush\\
        \hline
        DataGen & \Checkmark & un-braided and a few braided & 231 & \Checkmark\\
        \hline
        \dataset & \Checkmark & 1,544 un-braided and 852 braided & 2,396 & 82,682\\
    
  \hline
  \end{tabular}
  \caption{Comparisons of current 3D hair datasets.}
 \label{tab:compare_dataset}
 \end{center}
\end{table*}

%% file: fig/dataset_gallery.tex
\begin{figure*}[t]
\centering
\includegraphics[width=0.95\textwidth]{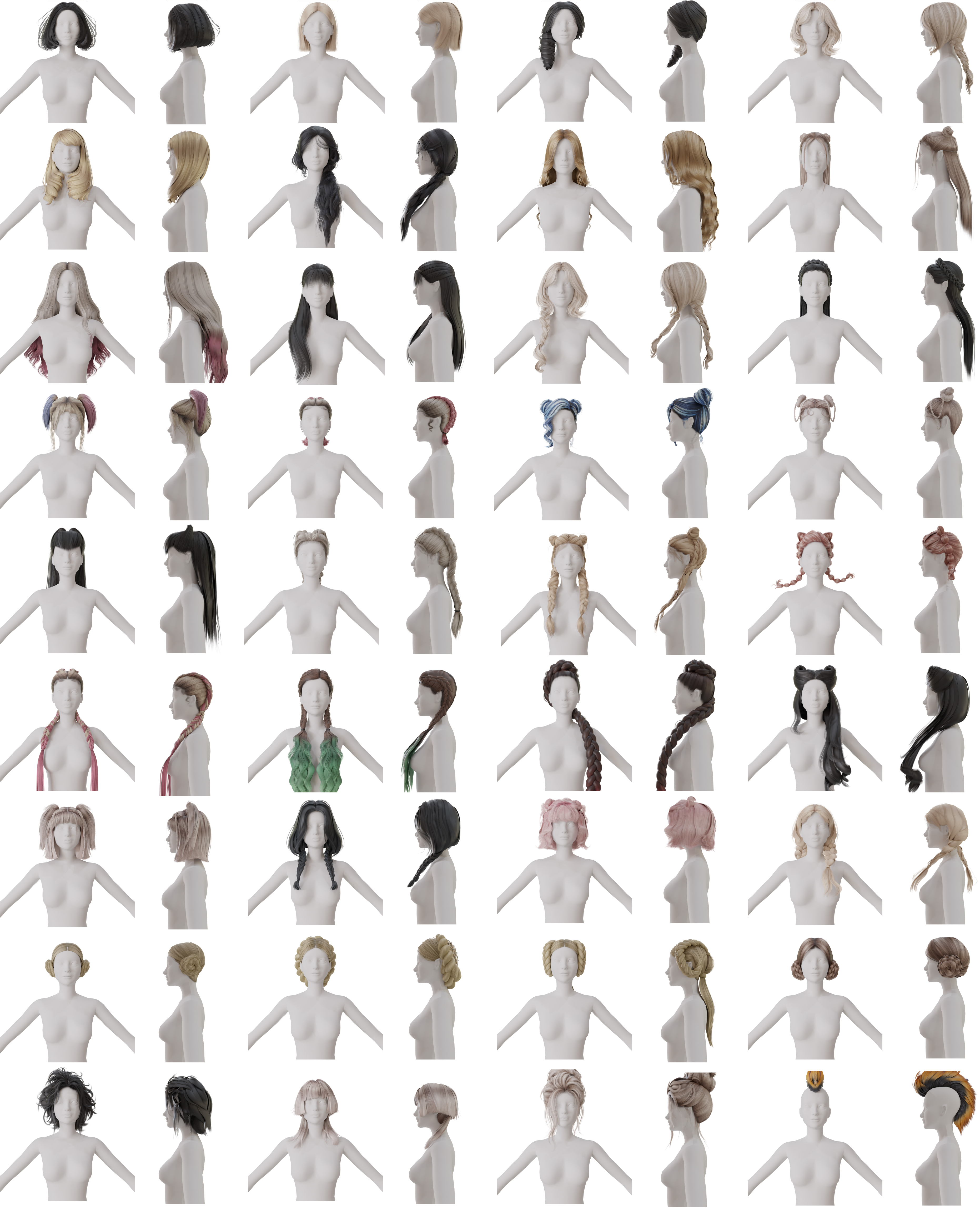}
\caption{\dataset dataset gallery of highly diverse hairstyles in styles. We show the frontal and side view for each hairstyle.}
\label{fig:dataset_shape}
\end{figure*}

\begin{figure*}[t]
\centering
\includegraphics[width=0.95\textwidth]{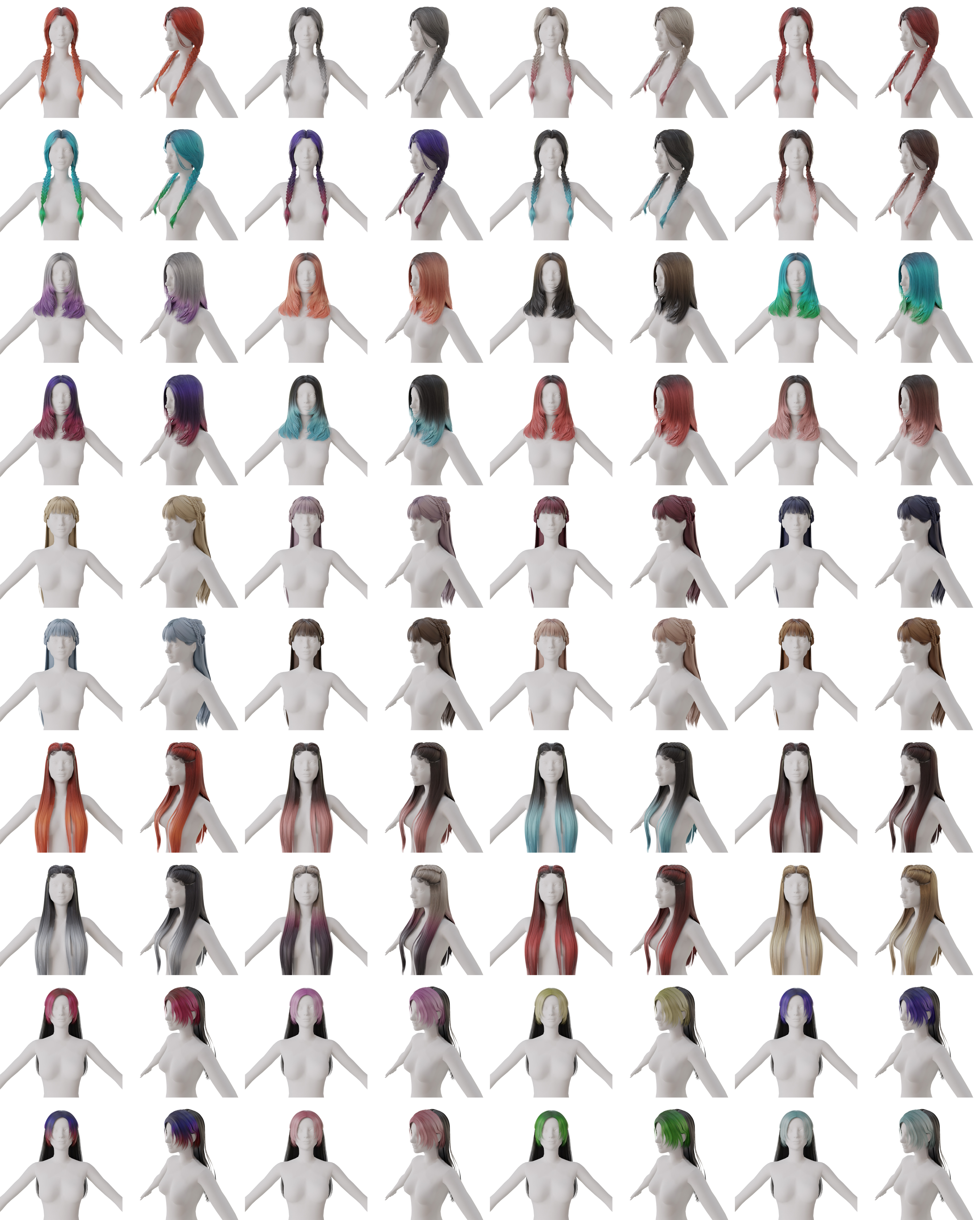}
\caption{\dataset dataset gallery of highly diverse hairstyles in texture. Each two rows show the different hair textures in two rendering poses.}
\label{fig:dataset_tex}
\end{figure*}

\begin{figure*}[t]
\centering
\includegraphics[width=0.98\textwidth]{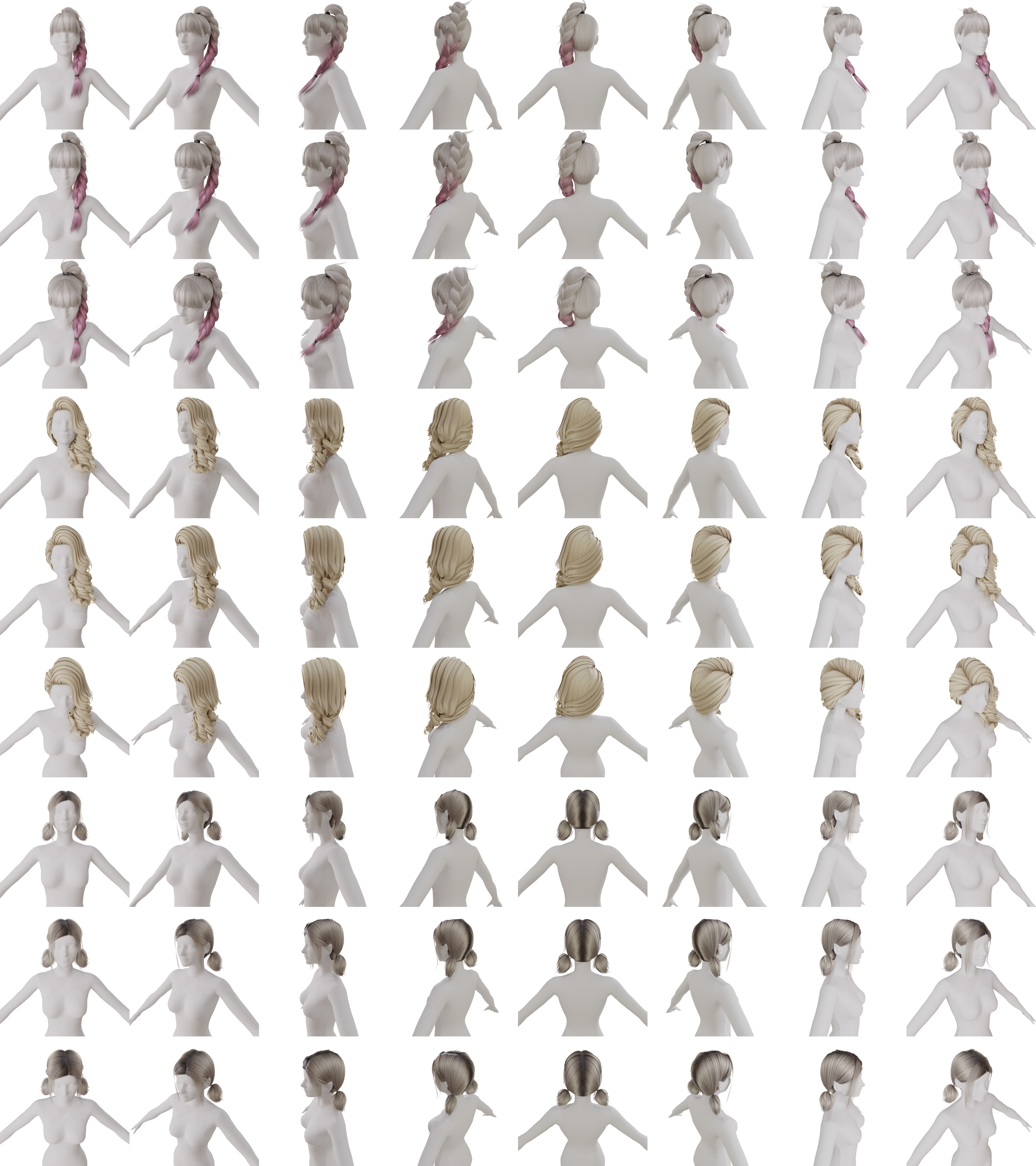}
\caption{\dataset dataset gallery of highly diverse renderings camera poses. Each three rows gives the rendering of one hairstyle with different azimuth and elevation angle}
\label{fig:dataset_view}
\end{figure*}

%% file: fig/ablation_vgg.tex
\begin{figure*}[t]
\centering
\includegraphics[width=0.85\textwidth]{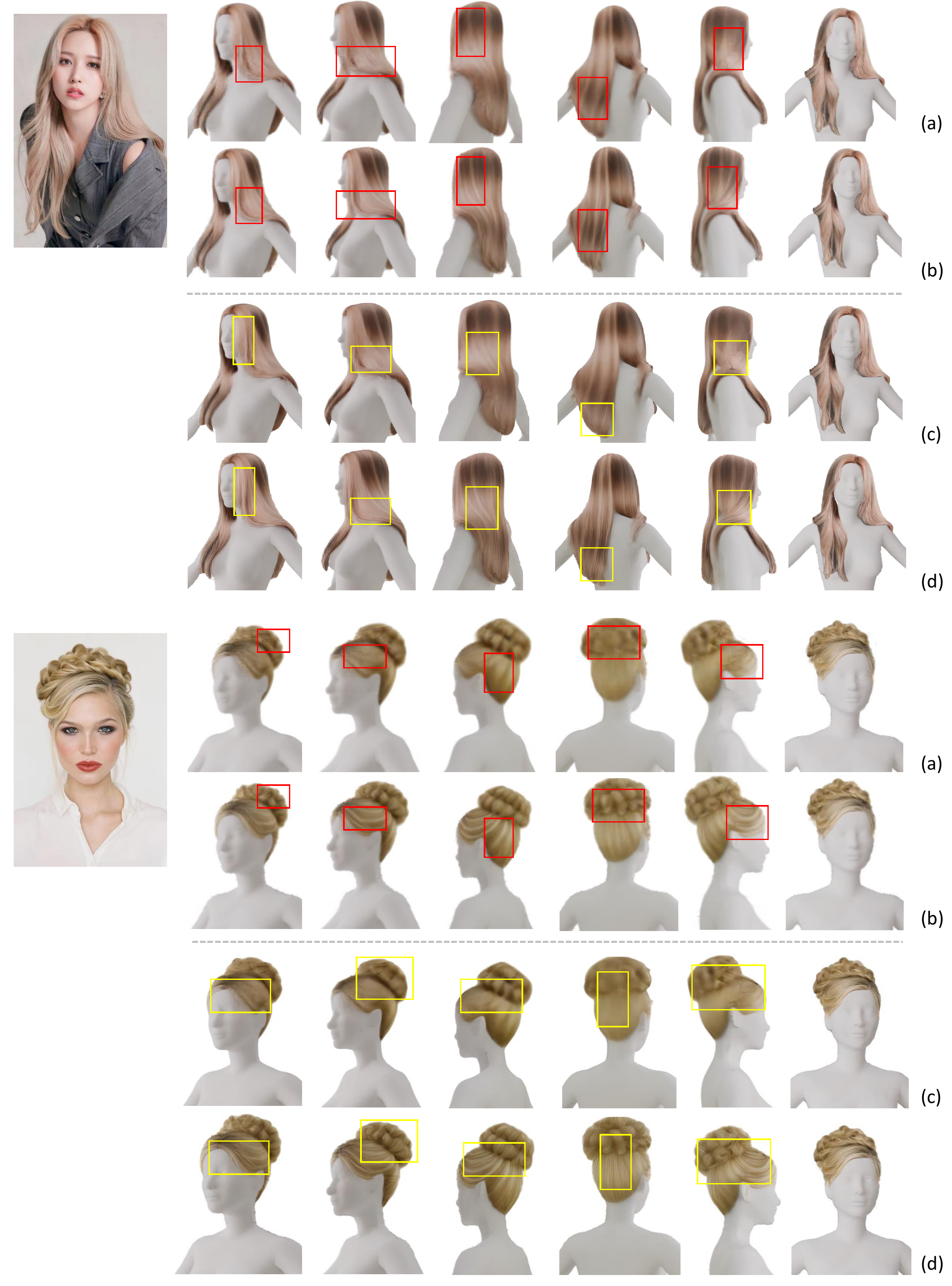}

\caption{Ablation study on perceptual loss. (a) Rendering results of refined Gaussian \refinedgs w/o perceptual loss. (b) Rendering results of refined Gaussian \refinedgs w/ perceptual loss. (c) Rendering results of enhanced final Gaussian \enhancedgs w/o perceptual loss. (d) Rendering results of enhanced final Gaussian \enhancedgs w/ perceptual loss.}
\label{fig:ablation_vgg}
\end{figure*}

%% file: table/ablation_vgg.tex
\begin{table}
 \begin{center}
  \begin{tabular}{l|c c c c}
   \hline
   Method & \refinedgs w/o $p$ & \refinedgs w/ $p$ & \enhancedgs w/o $p$ & \enhancedgs w/ $p$\\
        \hline
        $L_1$ $\downarrow$ & 0.01276 & 0.01249 & 0.01291 & 0.01236 \\
        $Perceptual$ $\downarrow$ & 2.678 & 2.403 & 2.629 & 2.352 \\
        $PSNR$ $\uparrow$ & 28.69 & 28.99 & 29.40 & 29.60  \\
   \hline
  \end{tabular}
  \caption{Ablation study on perceptual loss.}
 \label{tab:ablation_vgg_tab}
 \end{center}
\end{table} 

%% file: fig/compare_dreamgs.tex
\begin{figure*}[t]
\centering
\includegraphics[width=0.98\textwidth]{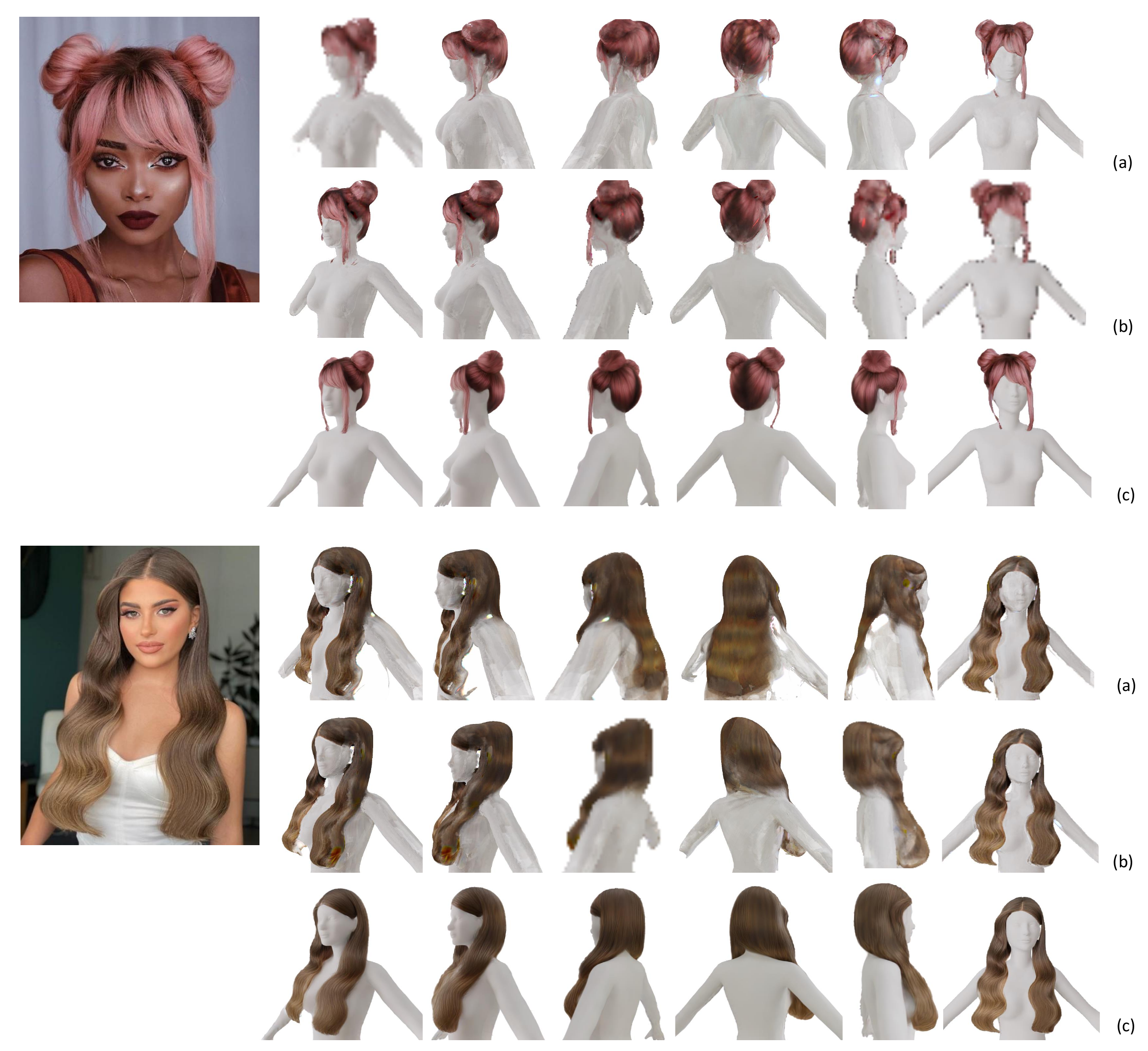}
\caption{Comparisons between (a) DreamGaussian~\cite{tang2023dreamgaussian}, (b) DreamGaussian with our prior \synthesizer and (c) our method.}
\label{fig:compare_dreamgs}
\end{figure*}

%% file: table/compare_dreamgs.tex
\begin{table}
 \begin{center}
  \begin{tabular}{l|c c c c}
   \hline
   Method& $L_1$ $\downarrow$ & $Perceptual$ $\downarrow$ & $PSNR$ $\uparrow$\\     
        \hline
        DreamGaussian & 0.02832 & 4.919 & 23.36 \\
        DreamGaussian-ourPrior & 0.01988 & 4.170 & 25.53 \\
        Ours& 0.01236 & 2.352 & 29.60  \\
   \hline
  \end{tabular}
  \caption{Comparisons with DreamGaussian~\cite{tang2023dreamgaussian}.}
 \label{tab:compare_dreamgs_tab}
 \end{center}
\end{table} 

%% file: fig/more_results1.tex
\begin{figure*}[t]
\centering
\includegraphics[width=1.0\textwidth]{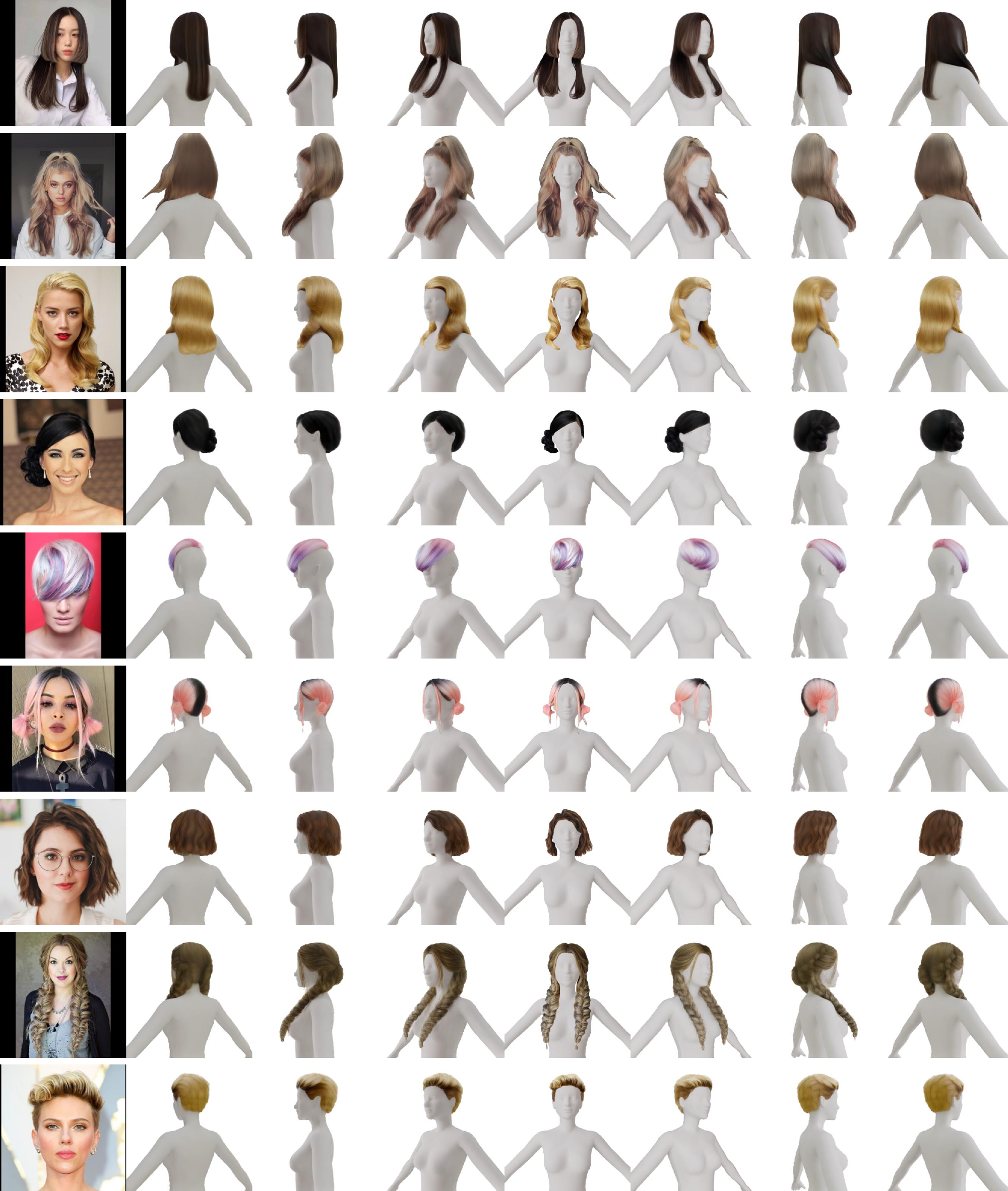}
\caption{More results on in-the-wild images. We render our reconstructed 3D hair from 7 views for illustration.}
\label{fig:more_results1}
\end{figure*}

%% file: fig/more_results2.tex
\begin{figure*}[t]
\centering
\includegraphics[width=1.0\textwidth]{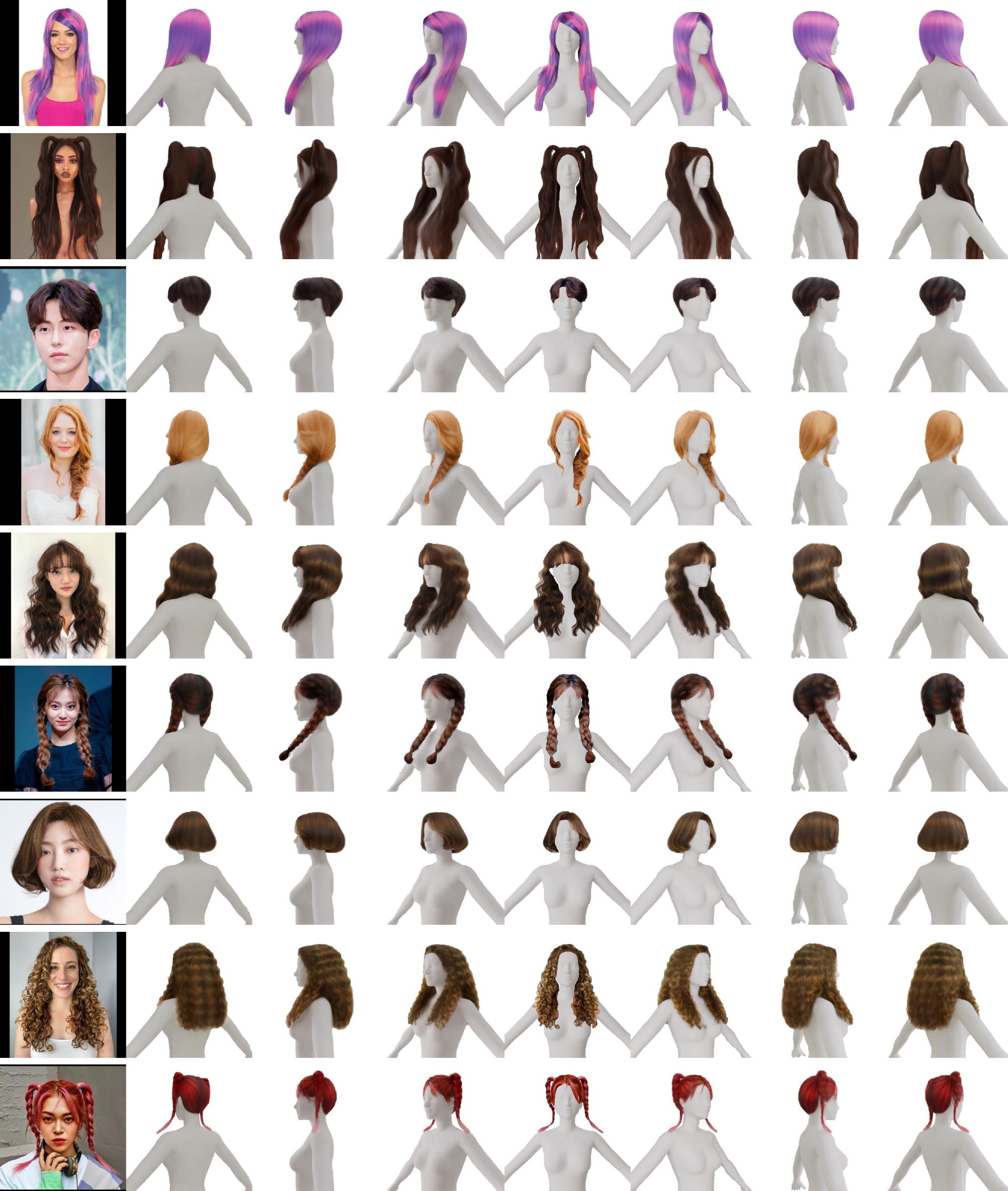}
\caption{More results on in-the-wild images. We render our reconstructed 3D hair from 7 views for illustration.}
\label{fig:more_results2}
\end{figure*}

%% file: main.bbl

\begin{thebibliography}{46}


\ifx \showCODEN    \undefined \def \showCODEN     #1{\unskip}     \fi
\ifx \showDOI      \undefined \def \showDOI       #1{#1}\fi
\ifx \showISBNx    \undefined \def \showISBNx     #1{\unskip}     \fi
\ifx \showISBNxiii \undefined \def \showISBNxiii  #1{\unskip}     \fi
\ifx \showISSN     \undefined \def \showISSN      #1{\unskip}     \fi
\ifx \showLCCN     \undefined \def \showLCCN      #1{\unskip}     \fi
\ifx \shownote     \undefined \def \shownote      #1{#1}          \fi
\ifx \showarticletitle \undefined \def \showarticletitle #1{#1}   \fi
\ifx \showURL      \undefined \def \showURL       {\relax}        \fi
\providecommand\bibfield[2]{#2}
\providecommand\bibinfo[2]{#2}
\providecommand\natexlab[1]{#1}
\providecommand\showeprint[2][]{arXiv:#2}

\bibitem[An et~al\mbox{.}(2023)]%
        {an2023panohead}
\bibfield{author}{\bibinfo{person}{Sizhe An}, \bibinfo{person}{Hongyi Xu}, \bibinfo{person}{Yichun Shi}, \bibinfo{person}{Guoxian Song}, \bibinfo{person}{Umit~Y Ogras}, {and} \bibinfo{person}{Linjie Luo}.} \bibinfo{year}{2023}\natexlab{}.
\newblock \showarticletitle{Panohead: Geometry-aware 3d full-head synthesis in 360deg}. In \bibinfo{booktitle}{\emph{Proceedings of the IEEE/CVF Conference on Computer Vision and Pattern Recognition}}. \bibinfo{pages}{20950--20959}.
\newblock


\bibitem[Chai et~al\mbox{.}(2015)]%
        {chai2015high}
\bibfield{author}{\bibinfo{person}{Menglei Chai}, \bibinfo{person}{Linjie Luo}, \bibinfo{person}{Kalyan Sunkavalli}, \bibinfo{person}{Nathan Carr}, \bibinfo{person}{Sunil Hadap}, {and} \bibinfo{person}{Kun Zhou}.} \bibinfo{year}{2015}\natexlab{}.
\newblock \showarticletitle{High-quality hair modeling from a single portrait photo.}
\newblock \bibinfo{journal}{\emph{ACM Trans. Graph.}} \bibinfo{volume}{34}, \bibinfo{number}{6} (\bibinfo{year}{2015}), \bibinfo{pages}{204--1}.
\newblock


\bibitem[Chai et~al\mbox{.}(2016)]%
        {chai2016autohair}
\bibfield{author}{\bibinfo{person}{Menglei Chai}, \bibinfo{person}{Tianjia Shao}, \bibinfo{person}{Hongzhi Wu}, \bibinfo{person}{Yanlin Weng}, {and} \bibinfo{person}{Kun Zhou}.} \bibinfo{year}{2016}\natexlab{}.
\newblock \showarticletitle{Autohair: Fully automatic hair modeling from a single image}.
\newblock \bibinfo{journal}{\emph{ACM Transactions on Graphics}} \bibinfo{volume}{35}, \bibinfo{number}{4} (\bibinfo{year}{2016}).
\newblock


\bibitem[Chai et~al\mbox{.}(2013)]%
        {chai2013dynamic}
\bibfield{author}{\bibinfo{person}{Menglei Chai}, \bibinfo{person}{Lvdi Wang}, \bibinfo{person}{Yanlin Weng}, \bibinfo{person}{Xiaogang Jin}, {and} \bibinfo{person}{Kun Zhou}.} \bibinfo{year}{2013}\natexlab{}.
\newblock \showarticletitle{Dynamic hair manipulation in images and videos}.
\newblock \bibinfo{journal}{\emph{ACM Transactions on Graphics (TOG)}} \bibinfo{volume}{32}, \bibinfo{number}{4} (\bibinfo{year}{2013}), \bibinfo{pages}{1--8}.
\newblock


\bibitem[Chai et~al\mbox{.}(2012)]%
        {chai2012single}
\bibfield{author}{\bibinfo{person}{Menglei Chai}, \bibinfo{person}{Lvdi Wang}, \bibinfo{person}{Yanlin Weng}, \bibinfo{person}{Yizhou Yu}, \bibinfo{person}{Baining Guo}, {and} \bibinfo{person}{Kun Zhou}.} \bibinfo{year}{2012}\natexlab{}.
\newblock \showarticletitle{Single-view hair modeling for portrait manipulation}.
\newblock \bibinfo{journal}{\emph{ACM Transactions on Graphics (TOG)}} \bibinfo{volume}{31}, \bibinfo{number}{4} (\bibinfo{year}{2012}), \bibinfo{pages}{1--8}.
\newblock


\bibitem[Deitke et~al\mbox{.}(2023)]%
        {deitke2023objaverse}
\bibfield{author}{\bibinfo{person}{Matt Deitke}, \bibinfo{person}{Dustin Schwenk}, \bibinfo{person}{Jordi Salvador}, \bibinfo{person}{Luca Weihs}, \bibinfo{person}{Oscar Michel}, \bibinfo{person}{Eli VanderBilt}, \bibinfo{person}{Ludwig Schmidt}, \bibinfo{person}{Kiana Ehsani}, \bibinfo{person}{Aniruddha Kembhavi}, {and} \bibinfo{person}{Ali Farhadi}.} \bibinfo{year}{2023}\natexlab{}.
\newblock \showarticletitle{Objaverse: A universe of annotated 3d objects}. In \bibinfo{booktitle}{\emph{Proceedings of the IEEE/CVF Conference on Computer Vision and Pattern Recognition}}. \bibinfo{pages}{13142--13153}.
\newblock


\bibitem[Guo et~al\mbox{.}(2020)]%
        {guo2020towards}
\bibfield{author}{\bibinfo{person}{Jianzhu Guo}, \bibinfo{person}{Xiangyu Zhu}, \bibinfo{person}{Yang Yang}, \bibinfo{person}{Fan Yang}, \bibinfo{person}{Zhen Lei}, {and} \bibinfo{person}{Stan~Z Li}.} \bibinfo{year}{2020}\natexlab{}.
\newblock \showarticletitle{Towards fast, accurate and stable 3d dense face alignment}. In \bibinfo{booktitle}{\emph{European Conference on Computer Vision}}. Springer, \bibinfo{pages}{152--168}.
\newblock


\bibitem[Hu et~al\mbox{.}(2014a)]%
        {hu2014robust}
\bibfield{author}{\bibinfo{person}{Liwen Hu}, \bibinfo{person}{Chongyang Ma}, \bibinfo{person}{Linjie Luo}, {and} \bibinfo{person}{Hao Li}.} \bibinfo{year}{2014}\natexlab{a}.
\newblock \showarticletitle{Robust hair capture using simulated examples}.
\newblock \bibinfo{journal}{\emph{ACM Transactions on Graphics (TOG)}} \bibinfo{volume}{33}, \bibinfo{number}{4} (\bibinfo{year}{2014}), \bibinfo{pages}{1--10}.
\newblock


\bibitem[Hu et~al\mbox{.}(2015)]%
        {hu2015single}
\bibfield{author}{\bibinfo{person}{Liwen Hu}, \bibinfo{person}{Chongyang Ma}, \bibinfo{person}{Linjie Luo}, {and} \bibinfo{person}{Hao Li}.} \bibinfo{year}{2015}\natexlab{}.
\newblock \showarticletitle{Single-view hair modeling using a hairstyle database}.
\newblock \bibinfo{journal}{\emph{ACM Transactions on Graphics (ToG)}} \bibinfo{volume}{34}, \bibinfo{number}{4} (\bibinfo{year}{2015}), \bibinfo{pages}{1--9}.
\newblock


\bibitem[Hu et~al\mbox{.}(2014b)]%
        {hu2014capturing}
\bibfield{author}{\bibinfo{person}{Liwen Hu}, \bibinfo{person}{Chongyang Ma}, \bibinfo{person}{Linjie Luo}, \bibinfo{person}{Li-Yi Wei}, {and} \bibinfo{person}{Hao Li}.} \bibinfo{year}{2014}\natexlab{b}.
\newblock \showarticletitle{Capturing braided hairstyles}.
\newblock \bibinfo{journal}{\emph{ACM Transactions on Graphics (TOG)}} \bibinfo{volume}{33}, \bibinfo{number}{6} (\bibinfo{year}{2014}), \bibinfo{pages}{1--9}.
\newblock


\bibitem[Hu et~al\mbox{.}(2017)]%
        {hu2017avatar}
\bibfield{author}{\bibinfo{person}{Liwen Hu}, \bibinfo{person}{Shunsuke Saito}, \bibinfo{person}{Lingyu Wei}, \bibinfo{person}{Koki Nagano}, \bibinfo{person}{Jaewoo Seo}, \bibinfo{person}{Jens Fursund}, \bibinfo{person}{Iman Sadeghi}, \bibinfo{person}{Carrie Sun}, \bibinfo{person}{Yen-Chun Chen}, {and} \bibinfo{person}{Hao Li}.} \bibinfo{year}{2017}\natexlab{}.
\newblock \showarticletitle{Avatar digitization from a single image for real-time rendering}.
\newblock \bibinfo{journal}{\emph{ACM Transactions on Graphics (ToG)}} \bibinfo{volume}{36}, \bibinfo{number}{6} (\bibinfo{year}{2017}), \bibinfo{pages}{1--14}.
\newblock


\bibitem[Johnson et~al\mbox{.}(2016)]%
        {johnson2016perceptual}
\bibfield{author}{\bibinfo{person}{Justin Johnson}, \bibinfo{person}{Alexandre Alahi}, {and} \bibinfo{person}{Li Fei-Fei}.} \bibinfo{year}{2016}\natexlab{}.
\newblock \showarticletitle{Perceptual losses for real-time style transfer and super-resolution}. In \bibinfo{booktitle}{\emph{Computer Vision--ECCV 2016: 14th European Conference, Amsterdam, The Netherlands, October 11-14, 2016, Proceedings, Part II 14}}. Springer, \bibinfo{pages}{694--711}.
\newblock


\bibitem[Kerbl et~al\mbox{.}(2023)]%
        {kerbl3Dgaussians}
\bibfield{author}{\bibinfo{person}{Bernhard Kerbl}, \bibinfo{person}{Georgios Kopanas}, \bibinfo{person}{Thomas Leimk{\"u}hler}, {and} \bibinfo{person}{George Drettakis}.} \bibinfo{year}{2023}\natexlab{}.
\newblock \showarticletitle{3D Gaussian Splatting for Real-Time Radiance Field Rendering}.
\newblock \bibinfo{journal}{\emph{ACM Transactions on Graphics}} \bibinfo{volume}{42}, \bibinfo{number}{4} (\bibinfo{date}{July} \bibinfo{year}{2023}).
\newblock
\urldef\tempurl%
\url{https://repo-sam.inria.fr/fungraph/3d-gaussian-splatting/}
\showURL{%
\tempurl}


\bibitem[Kim et~al\mbox{.}(2021)]%
        {kim2021k}
\bibfield{author}{\bibinfo{person}{Taewoo Kim}, \bibinfo{person}{Chaeyeon Chung}, \bibinfo{person}{Sunghyun Park}, \bibinfo{person}{Gyojung Gu}, \bibinfo{person}{Keonmin Nam}, \bibinfo{person}{Wonzo Choe}, \bibinfo{person}{Jaesung Lee}, {and} \bibinfo{person}{Jaegul Choo}.} \bibinfo{year}{2021}\natexlab{}.
\newblock \showarticletitle{K-hairstyle: A large-scale korean hairstyle dataset for virtual hair editing and hairstyle classification}. In \bibinfo{booktitle}{\emph{2021 IEEE International Conference on Image Processing (ICIP)}}. IEEE, \bibinfo{pages}{1299--1303}.
\newblock


\bibitem[Kirillov et~al\mbox{.}(2023)]%
        {kirillov2023segany}
\bibfield{author}{\bibinfo{person}{Alexander Kirillov}, \bibinfo{person}{Eric Mintun}, \bibinfo{person}{Nikhila Ravi}, \bibinfo{person}{Hanzi Mao}, \bibinfo{person}{Chloe Rolland}, \bibinfo{person}{Laura Gustafson}, \bibinfo{person}{Tete Xiao}, \bibinfo{person}{Spencer Whitehead}, \bibinfo{person}{Alexander~C. Berg}, \bibinfo{person}{Wan-Yen Lo}, \bibinfo{person}{Piotr Doll{\'a}r}, {and} \bibinfo{person}{Ross Girshick}.} \bibinfo{year}{2023}\natexlab{}.
\newblock \showarticletitle{Segment Anything}.
\newblock \bibinfo{journal}{\emph{arXiv:2304.02643}} (\bibinfo{year}{2023}).
\newblock


\bibitem[Kuang et~al\mbox{.}(2022)]%
        {kuang2022deepmvshair}
\bibfield{author}{\bibinfo{person}{Zhiyi Kuang}, \bibinfo{person}{Yiyang Chen}, \bibinfo{person}{Hongbo Fu}, \bibinfo{person}{Kun Zhou}, {and} \bibinfo{person}{Youyi Zheng}.} \bibinfo{year}{2022}\natexlab{}.
\newblock \showarticletitle{Deepmvshair: Deep hair modeling from sparse views}. In \bibinfo{booktitle}{\emph{SIGGRAPH Asia 2022 Conference Papers}}. \bibinfo{pages}{1--8}.
\newblock


\bibitem[Liang et~al\mbox{.}(2018)]%
        {liang2018video}
\bibfield{author}{\bibinfo{person}{Shu Liang}, \bibinfo{person}{Xiufeng Huang}, \bibinfo{person}{Xianyu Meng}, \bibinfo{person}{Kunyao Chen}, \bibinfo{person}{Linda~G Shapiro}, {and} \bibinfo{person}{Ira Kemelmacher-Shlizerman}.} \bibinfo{year}{2018}\natexlab{}.
\newblock \showarticletitle{Video to fully automatic 3d hair model}.
\newblock \bibinfo{journal}{\emph{ACM Transactions on Graphics (TOG)}} \bibinfo{volume}{37}, \bibinfo{number}{6} (\bibinfo{year}{2018}), \bibinfo{pages}{1--14}.
\newblock


\bibitem[Liu et~al\mbox{.}(2023a)]%
        {liu2023one1}
\bibfield{author}{\bibinfo{person}{Minghua Liu}, \bibinfo{person}{Ruoxi Shi}, \bibinfo{person}{Linghao Chen}, \bibinfo{person}{Zhuoyang Zhang}, \bibinfo{person}{Chao Xu}, \bibinfo{person}{Xinyue Wei}, \bibinfo{person}{Hansheng Chen}, \bibinfo{person}{Chong Zeng}, \bibinfo{person}{Jiayuan Gu}, {and} \bibinfo{person}{Hao Su}.} \bibinfo{year}{2023}\natexlab{a}.
\newblock \showarticletitle{One-2-3-45++: Fast single image to 3d objects with consistent multi-view generation and 3d diffusion}.
\newblock \bibinfo{journal}{\emph{arXiv preprint arXiv:2311.07885}} (\bibinfo{year}{2023}).
\newblock


\bibitem[Liu et~al\mbox{.}(2023c)]%
        {liu2023one}
\bibfield{author}{\bibinfo{person}{Minghua Liu}, \bibinfo{person}{Chao Xu}, \bibinfo{person}{Haian Jin}, \bibinfo{person}{Linghao Chen}, \bibinfo{person}{Zexiang Xu}, \bibinfo{person}{Hao Su}, {et~al\mbox{.}}} \bibinfo{year}{2023}\natexlab{c}.
\newblock \showarticletitle{One-2-3-45: Any single image to 3d mesh in 45 seconds without per-shape optimization}.
\newblock \bibinfo{journal}{\emph{arXiv preprint arXiv:2306.16928}} (\bibinfo{year}{2023}).
\newblock


\bibitem[Liu et~al\mbox{.}(2023b)]%
        {liu2023zero}
\bibfield{author}{\bibinfo{person}{Ruoshi Liu}, \bibinfo{person}{Rundi Wu}, \bibinfo{person}{Basile Van~Hoorick}, \bibinfo{person}{Pavel Tokmakov}, \bibinfo{person}{Sergey Zakharov}, {and} \bibinfo{person}{Carl Vondrick}.} \bibinfo{year}{2023}\natexlab{b}.
\newblock \showarticletitle{Zero-1-to-3: Zero-shot one image to 3d object}. In \bibinfo{booktitle}{\emph{Proceedings of the IEEE/CVF International Conference on Computer Vision}}. \bibinfo{pages}{9298--9309}.
\newblock


\bibitem[Liu et~al\mbox{.}(2023d)]%
        {liu2023grounding}
\bibfield{author}{\bibinfo{person}{Shilong Liu}, \bibinfo{person}{Zhaoyang Zeng}, \bibinfo{person}{Tianhe Ren}, \bibinfo{person}{Feng Li}, \bibinfo{person}{Hao Zhang}, \bibinfo{person}{Jie Yang}, \bibinfo{person}{Chunyuan Li}, \bibinfo{person}{Jianwei Yang}, \bibinfo{person}{Hang Su}, \bibinfo{person}{Jun Zhu}, {et~al\mbox{.}}} \bibinfo{year}{2023}\natexlab{d}.
\newblock \showarticletitle{Grounding dino: Marrying dino with grounded pre-training for open-set object detection}.
\newblock \bibinfo{journal}{\emph{arXiv preprint arXiv:2303.05499}} (\bibinfo{year}{2023}).
\newblock


\bibitem[Luo et~al\mbox{.}(2024)]%
        {luo2024gaussianhair}
\bibfield{author}{\bibinfo{person}{Haimin Luo}, \bibinfo{person}{Min Ouyang}, \bibinfo{person}{Zijun Zhao}, \bibinfo{person}{Suyi Jiang}, \bibinfo{person}{Longwen Zhang}, \bibinfo{person}{Qixuan Zhang}, \bibinfo{person}{Wei Yang}, \bibinfo{person}{Lan Xu}, {and} \bibinfo{person}{Jingyi Yu}.} \bibinfo{year}{2024}\natexlab{}.
\newblock \showarticletitle{GaussianHair: Hair Modeling and Rendering with Light-aware Gaussians}.
\newblock \bibinfo{journal}{\emph{arXiv preprint arXiv:2402.10483}} (\bibinfo{year}{2024}).
\newblock


\bibitem[Luo et~al\mbox{.}(2012)]%
        {luo2012multi}
\bibfield{author}{\bibinfo{person}{Linjie Luo}, \bibinfo{person}{Hao Li}, \bibinfo{person}{Sylvain Paris}, \bibinfo{person}{Thibaut Weise}, \bibinfo{person}{Mark Pauly}, {and} \bibinfo{person}{Szymon Rusinkiewicz}.} \bibinfo{year}{2012}\natexlab{}.
\newblock \showarticletitle{Multi-view hair capture using orientation fields}. In \bibinfo{booktitle}{\emph{2012 IEEE Conference on Computer Vision and Pattern Recognition}}. IEEE, \bibinfo{pages}{1490--1497}.
\newblock


\bibitem[Luo et~al\mbox{.}(2013a)]%
        {luo2013structure}
\bibfield{author}{\bibinfo{person}{Linjie Luo}, \bibinfo{person}{Hao Li}, {and} \bibinfo{person}{Szymon Rusinkiewicz}.} \bibinfo{year}{2013}\natexlab{a}.
\newblock \showarticletitle{Structure-aware hair capture}.
\newblock \bibinfo{journal}{\emph{ACM Transactions on Graphics (TOG)}} \bibinfo{volume}{32}, \bibinfo{number}{4} (\bibinfo{year}{2013}), \bibinfo{pages}{1--12}.
\newblock


\bibitem[Luo et~al\mbox{.}(2013b)]%
        {luo2013wide}
\bibfield{author}{\bibinfo{person}{Linjie Luo}, \bibinfo{person}{Cha Zhang}, \bibinfo{person}{Zhengyou Zhang}, {and} \bibinfo{person}{Szymon Rusinkiewicz}.} \bibinfo{year}{2013}\natexlab{b}.
\newblock \showarticletitle{Wide-baseline hair capture using strand-based refinement}. In \bibinfo{booktitle}{\emph{Proceedings of the IEEE Conference on Computer Vision and Pattern Recognition}}. \bibinfo{pages}{265--272}.
\newblock


\bibitem[Meng et~al\mbox{.}(2022)]%
        {meng2022sdedit}
\bibfield{author}{\bibinfo{person}{Chenlin Meng}, \bibinfo{person}{Yutong He}, \bibinfo{person}{Yang Song}, \bibinfo{person}{Jiaming Song}, \bibinfo{person}{Jiajun Wu}, \bibinfo{person}{Jun-Yan Zhu}, {and} \bibinfo{person}{Stefano Ermon}.} \bibinfo{year}{2022}\natexlab{}.
\newblock \showarticletitle{{SDE}dit: Guided Image Synthesis and Editing with Stochastic Differential Equations}. In \bibinfo{booktitle}{\emph{International Conference on Learning Representations}}.
\newblock


\bibitem[Nam et~al\mbox{.}(2019)]%
        {nam2019strand}
\bibfield{author}{\bibinfo{person}{Giljoo Nam}, \bibinfo{person}{Chenglei Wu}, \bibinfo{person}{Min~H Kim}, {and} \bibinfo{person}{Yaser Sheikh}.} \bibinfo{year}{2019}\natexlab{}.
\newblock \showarticletitle{Strand-accurate multi-view hair capture}. In \bibinfo{booktitle}{\emph{Proceedings of the IEEE/CVF Conference on Computer Vision and Pattern Recognition}}. \bibinfo{pages}{155--164}.
\newblock


\bibitem[Poole et~al\mbox{.}(2022)]%
        {poole2022dreamfusion}
\bibfield{author}{\bibinfo{person}{Ben Poole}, \bibinfo{person}{Ajay Jain}, \bibinfo{person}{Jonathan~T Barron}, {and} \bibinfo{person}{Ben Mildenhall}.} \bibinfo{year}{2022}\natexlab{}.
\newblock \showarticletitle{Dreamfusion: Text-to-3d using 2d diffusion}.
\newblock \bibinfo{journal}{\emph{arXiv preprint arXiv:2209.14988}} (\bibinfo{year}{2022}).
\newblock


\bibitem[Radford et~al\mbox{.}(2021)]%
        {radford2021learning}
\bibfield{author}{\bibinfo{person}{Alec Radford}, \bibinfo{person}{Jong~Wook Kim}, \bibinfo{person}{Chris Hallacy}, \bibinfo{person}{Aditya Ramesh}, \bibinfo{person}{Gabriel Goh}, \bibinfo{person}{Sandhini Agarwal}, \bibinfo{person}{Girish Sastry}, \bibinfo{person}{Amanda Askell}, \bibinfo{person}{Pamela Mishkin}, \bibinfo{person}{Jack Clark}, {et~al\mbox{.}}} \bibinfo{year}{2021}\natexlab{}.
\newblock \showarticletitle{Learning transferable visual models from natural language supervision}. In \bibinfo{booktitle}{\emph{International conference on machine learning}}. PMLR, \bibinfo{pages}{8748--8763}.
\newblock


\bibitem[Ren et~al\mbox{.}(2024)]%
        {ren2024grounded}
\bibfield{author}{\bibinfo{person}{Tianhe Ren}, \bibinfo{person}{Shilong Liu}, \bibinfo{person}{Ailing Zeng}, \bibinfo{person}{Jing Lin}, \bibinfo{person}{Kunchang Li}, \bibinfo{person}{He Cao}, \bibinfo{person}{Jiayu Chen}, \bibinfo{person}{Xinyu Huang}, \bibinfo{person}{Yukang Chen}, \bibinfo{person}{Feng Yan}, \bibinfo{person}{Zhaoyang Zeng}, \bibinfo{person}{Hao Zhang}, \bibinfo{person}{Feng Li}, \bibinfo{person}{Jie Yang}, \bibinfo{person}{Hongyang Li}, \bibinfo{person}{Qing Jiang}, {and} \bibinfo{person}{Lei Zhang}.} \bibinfo{year}{2024}\natexlab{}.
\newblock \bibinfo{title}{Grounded SAM: Assembling Open-World Models for Diverse Visual Tasks}.
\newblock
\newblock
\showeprint[arxiv]{2401.14159}~[cs.CV]


\bibitem[Rosu et~al\mbox{.}(2022)]%
        {rosu2022neural}
\bibfield{author}{\bibinfo{person}{Radu~Alexandru Rosu}, \bibinfo{person}{Shunsuke Saito}, \bibinfo{person}{Ziyan Wang}, \bibinfo{person}{Chenglei Wu}, \bibinfo{person}{Sven Behnke}, {and} \bibinfo{person}{Giljoo Nam}.} \bibinfo{year}{2022}\natexlab{}.
\newblock \showarticletitle{Neural strands: Learning hair geometry and appearance from multi-view images}. In \bibinfo{booktitle}{\emph{European Conference on Computer Vision}}. Springer, \bibinfo{pages}{73--89}.
\newblock


\bibitem[Saito et~al\mbox{.}(2018)]%
        {saito20183d}
\bibfield{author}{\bibinfo{person}{Shunsuke Saito}, \bibinfo{person}{Liwen Hu}, \bibinfo{person}{Chongyang Ma}, \bibinfo{person}{Hikaru Ibayashi}, \bibinfo{person}{Linjie Luo}, {and} \bibinfo{person}{Hao Li}.} \bibinfo{year}{2018}\natexlab{}.
\newblock \showarticletitle{3D hair synthesis using volumetric variational autoencoders}.
\newblock \bibinfo{journal}{\emph{ACM Transactions on Graphics (TOG)}} \bibinfo{volume}{37}, \bibinfo{number}{6} (\bibinfo{year}{2018}), \bibinfo{pages}{1--12}.
\newblock


\bibitem[Shen et~al\mbox{.}(2023)]%
        {shen2023ct2hair}
\bibfield{author}{\bibinfo{person}{Yuefan Shen}, \bibinfo{person}{Shunsuke Saito}, \bibinfo{person}{Ziyan Wang}, \bibinfo{person}{Olivier Maury}, \bibinfo{person}{Chenglei Wu}, \bibinfo{person}{Jessica Hodgins}, \bibinfo{person}{Youyi Zheng}, {and} \bibinfo{person}{Giljoo Nam}.} \bibinfo{year}{2023}\natexlab{}.
\newblock \showarticletitle{CT2Hair: High-Fidelity 3D Hair Modeling using Computed Tomography}.
\newblock \bibinfo{journal}{\emph{ACM Transactions on Graphics (TOG)}} \bibinfo{volume}{42}, \bibinfo{number}{4} (\bibinfo{year}{2023}), \bibinfo{pages}{1--13}.
\newblock


\bibitem[Simonyan and Zisserman(2014)]%
        {simonyan2014very}
\bibfield{author}{\bibinfo{person}{Karen Simonyan} {and} \bibinfo{person}{Andrew Zisserman}.} \bibinfo{year}{2014}\natexlab{}.
\newblock \showarticletitle{Very deep convolutional networks for large-scale image recognition}.
\newblock \bibinfo{journal}{\emph{arXiv preprint arXiv:1409.1556}} (\bibinfo{year}{2014}).
\newblock


\bibitem[Sklyarova et~al\mbox{.}(2023)]%
        {sklyarova2023neural}
\bibfield{author}{\bibinfo{person}{Vanessa Sklyarova}, \bibinfo{person}{Jenya Chelishev}, \bibinfo{person}{Andreea Dogaru}, \bibinfo{person}{Igor Medvedev}, \bibinfo{person}{Victor Lempitsky}, {and} \bibinfo{person}{Egor Zakharov}.} \bibinfo{year}{2023}\natexlab{}.
\newblock \showarticletitle{Neural Haircut: Prior-Guided Strand-Based Hair Reconstruction}.
\newblock \bibinfo{journal}{\emph{arXiv preprint arXiv:2306.05872}} (\bibinfo{year}{2023}).
\newblock


\bibitem[Sun et~al\mbox{.}(2021)]%
        {sun2021single}
\bibfield{author}{\bibinfo{person}{Chao Sun}, \bibinfo{person}{Srinivasan Ramachandran}, \bibinfo{person}{Eric Paquette}, {and} \bibinfo{person}{Won-Sook Lee}.} \bibinfo{year}{2021}\natexlab{}.
\newblock \showarticletitle{Single-view procedural braided hair modeling through braid unit identification}.
\newblock \bibinfo{journal}{\emph{Computer Animation and Virtual Worlds}} \bibinfo{volume}{32}, \bibinfo{number}{3-4} (\bibinfo{year}{2021}), \bibinfo{pages}{e2007}.
\newblock


\bibitem[Tang et~al\mbox{.}(2023)]%
        {tang2023dreamgaussian}
\bibfield{author}{\bibinfo{person}{Jiaxiang Tang}, \bibinfo{person}{Jiawei Ren}, \bibinfo{person}{Hang Zhou}, \bibinfo{person}{Ziwei Liu}, {and} \bibinfo{person}{Gang Zeng}.} \bibinfo{year}{2023}\natexlab{}.
\newblock \showarticletitle{Dreamgaussian: Generative gaussian splatting for efficient 3d content creation}.
\newblock \bibinfo{journal}{\emph{arXiv preprint arXiv:2309.16653}} (\bibinfo{year}{2023}).
\newblock


\bibitem[Wu et~al\mbox{.}(2024)]%
        {wu2024monohair}
\bibfield{author}{\bibinfo{person}{Keyu Wu}, \bibinfo{person}{Lingchen Yang}, \bibinfo{person}{Zhiyi Kuang}, \bibinfo{person}{Yao Feng}, \bibinfo{person}{Xutao Han}, \bibinfo{person}{Yuefan Shen}, \bibinfo{person}{Hongbo Fu}, \bibinfo{person}{Kun Zhou}, {and} \bibinfo{person}{Youyi Zheng}.} \bibinfo{year}{2024}\natexlab{}.
\newblock \showarticletitle{MonoHair: High-Fidelity Hair Modeling from a Monocular Video}. In \bibinfo{booktitle}{\emph{Proceedings of the IEEE/CVF Conference on Computer Vision and Pattern Recognition}}. \bibinfo{pages}{24164--24173}.
\newblock


\bibitem[Wu et~al\mbox{.}(2022)]%
        {wu2022neuralhdhair}
\bibfield{author}{\bibinfo{person}{Keyu Wu}, \bibinfo{person}{Yifan Ye}, \bibinfo{person}{Lingchen Yang}, \bibinfo{person}{Hongbo Fu}, \bibinfo{person}{Kun Zhou}, {and} \bibinfo{person}{Youyi Zheng}.} \bibinfo{year}{2022}\natexlab{}.
\newblock \showarticletitle{Neuralhdhair: Automatic high-fidelity hair modeling from a single image using implicit neural representations}. In \bibinfo{booktitle}{\emph{Proceedings of the IEEE/CVF Conference on Computer Vision and Pattern Recognition}}. \bibinfo{pages}{1526--1535}.
\newblock


\bibitem[Yang et~al\mbox{.}(2019)]%
        {yang2019dynamic}
\bibfield{author}{\bibinfo{person}{Lingchen Yang}, \bibinfo{person}{Zefeng Shi}, \bibinfo{person}{Youyi Zheng}, {and} \bibinfo{person}{Kun Zhou}.} \bibinfo{year}{2019}\natexlab{}.
\newblock \showarticletitle{Dynamic hair modeling from monocular videos using deep neural networks}.
\newblock \bibinfo{journal}{\emph{ACM Transactions on Graphics (TOG)}} \bibinfo{volume}{38}, \bibinfo{number}{6} (\bibinfo{year}{2019}), \bibinfo{pages}{1--12}.
\newblock


\bibitem[Zhang et~al\mbox{.}(2017)]%
        {zhang2017data}
\bibfield{author}{\bibinfo{person}{Meng Zhang}, \bibinfo{person}{Menglei Chai}, \bibinfo{person}{Hongzhi Wu}, \bibinfo{person}{Hao Yang}, {and} \bibinfo{person}{Kun Zhou}.} \bibinfo{year}{2017}\natexlab{}.
\newblock \showarticletitle{A data-driven approach to four-view image-based hair modeling.}
\newblock \bibinfo{journal}{\emph{ACM Trans. Graph.}} \bibinfo{volume}{36}, \bibinfo{number}{4} (\bibinfo{year}{2017}), \bibinfo{pages}{156--1}.
\newblock


\bibitem[Zhang et~al\mbox{.}(2018)]%
        {zhang2018modeling}
\bibfield{author}{\bibinfo{person}{Meng Zhang}, \bibinfo{person}{Pan Wu}, \bibinfo{person}{Hongzhi Wu}, \bibinfo{person}{Yanlin Weng}, \bibinfo{person}{Youyi Zheng}, {and} \bibinfo{person}{Kun Zhou}.} \bibinfo{year}{2018}\natexlab{}.
\newblock \showarticletitle{Modeling hair from an rgb-d camera}.
\newblock \bibinfo{journal}{\emph{ACM Transactions on Graphics (TOG)}} \bibinfo{volume}{37}, \bibinfo{number}{6} (\bibinfo{year}{2018}), \bibinfo{pages}{1--10}.
\newblock


\bibitem[Zhang and Zheng(2019)]%
        {zhang2019hair}
\bibfield{author}{\bibinfo{person}{Meng Zhang} {and} \bibinfo{person}{Youyi Zheng}.} \bibinfo{year}{2019}\natexlab{}.
\newblock \showarticletitle{Hair-GAN: Recovering 3D hair structure from a single image using generative adversarial networks}.
\newblock \bibinfo{journal}{\emph{Visual Informatics}} \bibinfo{volume}{3}, \bibinfo{number}{2} (\bibinfo{year}{2019}), \bibinfo{pages}{102--112}.
\newblock


\bibitem[Zheng et~al\mbox{.}(2023)]%
        {zheng2023hairstep}
\bibfield{author}{\bibinfo{person}{Yujian Zheng}, \bibinfo{person}{Zirong Jin}, \bibinfo{person}{Moran Li}, \bibinfo{person}{Haibin Huang}, \bibinfo{person}{Chongyang Ma}, \bibinfo{person}{Shuguang Cui}, {and} \bibinfo{person}{Xiaoguang Han}.} \bibinfo{year}{2023}\natexlab{}.
\newblock \showarticletitle{Hairstep: Transfer synthetic to real using strand and depth maps for single-view 3d hair modeling}. In \bibinfo{booktitle}{\emph{Proceedings of the IEEE/CVF Conference on Computer Vision and Pattern Recognition}}. \bibinfo{pages}{12726--12735}.
\newblock


\bibitem[Zhou et~al\mbox{.}(2023)]%
        {zhou2023groomgen}
\bibfield{author}{\bibinfo{person}{Yuxiao Zhou}, \bibinfo{person}{Menglei Chai}, \bibinfo{person}{Alessandro Pepe}, \bibinfo{person}{Markus Gross}, {and} \bibinfo{person}{Thabo Beeler}.} \bibinfo{year}{2023}\natexlab{}.
\newblock \showarticletitle{GroomGen: A High-Quality Generative Hair Model Using Hierarchical Latent Representations}.
\newblock \bibinfo{journal}{\emph{ACM Transactions on Graphics (TOG)}} \bibinfo{volume}{42}, \bibinfo{number}{6} (\bibinfo{year}{2023}), \bibinfo{pages}{1--16}.
\newblock


\bibitem[Zhou et~al\mbox{.}(2018)]%
        {zhou2018hairnet}
\bibfield{author}{\bibinfo{person}{Yi Zhou}, \bibinfo{person}{Liwen Hu}, \bibinfo{person}{Jun Xing}, \bibinfo{person}{Weikai Chen}, \bibinfo{person}{Han-Wei Kung}, \bibinfo{person}{Xin Tong}, {and} \bibinfo{person}{Hao Li}.} \bibinfo{year}{2018}\natexlab{}.
\newblock \showarticletitle{Hairnet: Single-view hair reconstruction using convolutional neural networks}. In \bibinfo{booktitle}{\emph{Proceedings of the European Conference on Computer Vision (ECCV)}}. \bibinfo{pages}{235--251}.
\newblock


\end{thebibliography}
